\documentclass[runningheads]{llncs}

\usepackage{eccv}

\usepackage{eccvabbrv}

\usepackage{graphicx}
\usepackage{booktabs}
\usepackage{caption}
\usepackage{caption}

\usepackage{booktabs}
\usepackage{multirow}
\usepackage[table]{xcolor}

\usepackage[numbers]{natbib}
\usepackage{wrapfig}
\usepackage[accsupp]{axessibility}  % Improves PDF readability for those with disabilities.

\usepackage{hyperref}

\usepackage{orcidlink}

\begin{document}

\title{UniDrive-WM: Unified Understanding, Planning and Generation World Model for Autonomous Driving} 

\titlerunning{UniDrive-WM}

% Include the authors' ORCID for the camera-ready version, if at all possible.
\author{Zhexiao Xiong\inst{1,2}\thanks{Work was done during internship at Bosch Research North America.} \and
Xin Ye\inst{1} \and
Burhan Yaman\inst{1} \and
Sheng Cheng\inst{3} \and
Yiren Lu\inst{1,4} \and
Jingru Luo\inst{1} \and
Nathan Jacobs\inst{2} \and
Liu Ren\inst{1}}

\authorrunning{Z.~Xiong et al.}

\institute{Bosch Research North America \& Bosch Center for Artificial Intelligence (BCAI) \and
Washington University in St. Louis \and
Arizona State University \and
Case Western Reserve University}

\maketitle

\begin{center}

\includegraphics[width=\linewidth]{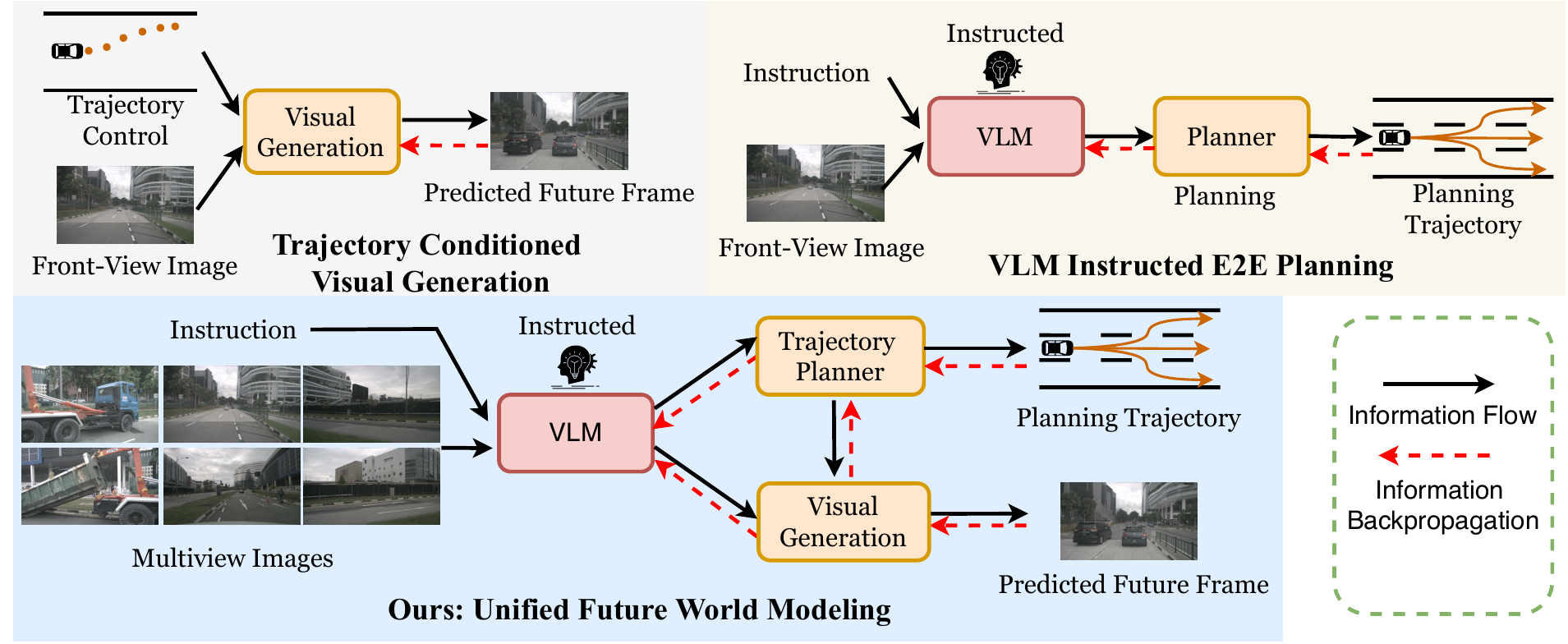}
\captionof{figure}{Top Left: trajectory-conditioned visual generation. Top Right: VLM-instructed E2E Planning. Bottom: our unified future world modeling method. Compared with conditioned future image generation models and VLM-instructed planning method, our framework establishes the connection among the reasoning, action and visual generation spaces via joint VLM-guided trajectory planner and future frame generation.}
\label{fig:teaser}
\end{center}

\begin{abstract}
World models have become central to autonomous driving, where accurate scene understanding and future prediction are crucial for safe control. Recent work has explored using vision–language models (VLMs) for planning, yet existing approaches typically treat perception, prediction, and planning as separate modules. We propose \textbf{UniDrive-WM}, a unified VLM-based world model that jointly performs driving-scene understanding, trajectory planning, and trajectory-conditioned future image generation within a single architecture. UniDrive-WM’s trajectory planner predicts a future trajectory, which conditions a VLM-based image generator to produce plausible future frames. These predictions provide additional supervisory signals that enhance scene understanding and iteratively refine trajectory generation. We further compare discrete and continuous output representations for future image prediction, analyzing their influence on downstream driving performance. Experiments on the challenging Bench2Drive benchmark show that UniDrive-WM produces high-fidelity future images and improves planning performance by 7.3\% in L2 trajectory error and 10.4\% in collision rate over the previous best method. These results demonstrate the advantages of tightly integrating VLM-driven reasoning, planning, and generative world modeling for autonomous driving. The project page is available at \url{https://unidrive-wm.github.io/UniDrive-WM}.
\end{abstract}

\section{Introduction}

Recent progress in multimodal large language models (MLLMs) has been propelled by the strong perception, reasoning, and instruction-following capabilities of vision–language models (VLMs). In parallel, visual generation has advanced along two complementary lines: autoregressive (AR) token prediction and diffusion-based continuous generation—enabling high-fidelity image synthesis across diverse tasks. Motivated by these trends, a growing body of \emph{unified models} seeks to couple understanding and generation within a single VLM, allowing the system to “think and generate” in a shared semantic space~\citep{li2025imagine,chern2025thinking,zeng2025futuresightdrive}.

In autonomous driving, world models \citep{zheng2025world4drive,liu2025towards,zheng2025world4drive} are typically defined as learned models that infer a latent representation of the current scene and predict future states conditioned on actions, ideally enabling robust planning and control. While several VLM-based approaches have been explored, many rely on text-only intermediates, first producing natural-language descriptions of future trajectories or scenes, and then invoking a separate stage to decode trajectories or render images. This pipeline introduces an information bottleneck: rich visual–geometric cues are abstracted into text, incurring inevitable loss and compounding errors across stages. Meanwhile, generative models can generate visually plausible frames, but they typically lack explicit state estimation and reasoning. In particular, they do not maintain structured representations of the scene and cannot reliably condition on multi-view cues or high-level instructions, which provides no differentiable bridge to the action space for verifiable planning. Consequently, they struggle to answer causal queries (e.g., “what if the pedestrian accelerates?”), to enforce safety constraints, or to propagate plan-consistent signals back into perception. In reality, a driver’s cognition is inherently joint: perceive the current scene, anticipate the future trajectory, and imagine the next visual scene. This motivates a unified formulation in which understanding, planning, and visual prediction are learned end-to-end within a single VLM framework, allowing information to flow bidirectionally between reasoning, action, and generation.

We present \textbf{UniDrive-WM}, a unified world model that \emph{jointly} performs scene understanding, trajectory planning, and future image generation within a VLM-centric architecture. As illustrated in Fig.~\ref{fig:teaser}, multi-view observations, temporal history, and perception cues (e.g., subject bounding boxes) are encoded and projected into an LLM reasoning space. A trajectory planner then produces a differentiable latent distribution over waypoints, bridging the reasoning (language–vision) space and the numeric action space. Conditioning on the predicted trajectory, UniDrive-WM further performs \emph{trajectory-conditioned} future image prediction via two complementary designs: (i) a discrete AR pathway that expands the visual codebook and detokenizes with MoVQGAN, and (ii) an AR+diffusion pathway that predicts continuous latent features with a flow-matching objective before pixel decoding. This design alleviates text-only bottlenecks and enables bidirectional coupling—both information flow and gradient flow—between reasoning, action, and generation. Extensive experiments on the challenging Bench2Drive and nuScenes datasets show that UniDrive-WM generates high-fidelity future image frames and outperforms counterpart methods on planning tasks across both open-loop and closed-loop metrics.

Our main contributions are summarized as follows:
\begin{itemize}
    \item We present \emph{UniDrive-WM}, a unified vision–language world model (VLM) that seamlessly integrates scene understanding, trajectory planning, and future image generation, enabling direct visual reasoning from spatio–temporal observations.
    
    \item We develop and analyze two complementary decoding paradigms for trajectory-conditioned future image prediction: a \emph{discrete} autoregressive (AR) pathway and a \emph{continuous} AR+diffusion pathway, revealing their respective advantages and trade-offs for autonomous driving.
    
    \item Extensive experiments demonstrate that our unified framework achieves high-fidelity, planning-conditioned visual generation and significantly improves both planning accuracy and perception performance on standard driving benchmarks, verifying the effectiveness of the proposed framework.
\end{itemize}

\section{Related Work}
\noindent\textbf{World Models for Autonomous Driving.}
The commonly accepted description of world models is understanding the current state and predicting the future state. For autonomous driving tasks, recent world-modeling-based methods aim to infer the ego status and dynamic environments from past observations to enable future planning and prediction, reducing human control in autonomous driving systems. Existing methods can be broadly categorized by their output modality: some focus on visual scene generation or urban view synthesis~\citep{hu2023gaia,gao2024vista,hassan2025gem,zhang2025epona, han2025extrapolated}, others emphasize future trajectory planning, including diffusion-based planning~\citep{fu2025orion,liao2025work,li2025drivevla,yu2026g2dp}, while several works address 3D scene reconstruction and simulation in the form of occupancy~\citep{wei2024occllama,wang2024occsora,liu2026occsim}, LiDAR~\citep{wu2024holodrive,zyrianov2025lidardm,wei2025lidardraft,liu2026towards}, or point clouds~\citep{zhang2023copilot4d,zhang2024bevworld}. However, most of these approaches focus on a single prediction modality. Recent world models have also been studied for visual navigation~\citep{dong2025unified,dong2026language} and driver-centric dynamics rollout~\citep{chi2026driver}. In contrast, our model jointly performs trajectory planning and future image generation within a unified world-model framework, coupling planned motion with predicted visual futures under a shared VLM backbone.

\noindent\textbf{Unified Image Understanding and Generation.}
Recent studies have emphasized the importance of unifying image understanding and image generation within a single framework. 
Early works such as Chameleon~\citep{team2024chameleon}, Show-o~\citep{xie2024showo}, Transfusion~\citep{zhou2024transfusion}, and Janus~\citep{wu2025janus} adopt discrete visual representations and treat visual synthesis as autoregressive token prediction, where images are quantized into discrete tokens analogous to text, enabling LLMs to interpret and generate visual content in a unified token space. 
More recent models, including Metamorph~\citep{tong2024metamorph} and the BLIP-3o family~\citep{chen2025blip3ofamilyfullyopen,chen2025blip3o}, integrate autoregressive generation with diffusion-based continuous decoding, often relying on additional visual encoders such as CLIP~\citep{radford2021learning} or SigLIP~\citep{zhai2023sigmoid} to enhance visual understanding. In contrast to these general-purpose unified models, our work advances this paradigm in the autonomous driving domain by jointly predicting future frames and planning trajectories within a single VLM-based world model.

\noindent\textbf{Vision–Language–Action Models.} 
Recently, large vision–language–action (VLA) models have emerged as powerful frameworks for embodied decision-making. 
Built upon pretrained multimodal large language models (MLLMs), these systems typically predict future actions through either discrete action decoders~\citep{kim2024openvla,cen2025worldvla,zhao2025cot} or continuous diffusion-based policy heads~\citep{li2025unified,he2025pre}. 
By leveraging large-scale and diverse training corpora, MLLMs provide strong semantic grounding and generalization capabilities for downstream embodied tasks. Recent driving-oriented VLA/action models further explore video-action priors and geometry-aware action modeling for autonomous driving~\citep{liu2026driveva,yao2026vlga}. In the autonomous driving setting, we treat ego-vehicle trajectories as the continuous action representation. Unlike standard manipulation or navigation environments, however, driving scenes are highly dynamic and visually complex, requiring the VLM to integrate richer contextual information—including temporal history, perception features, and multi-view observations to generate reliable and scene-consistent trajectory plans.

\section{Method}

\begin{figure*}[t!]
    \centering
    \includegraphics[width=\linewidth]{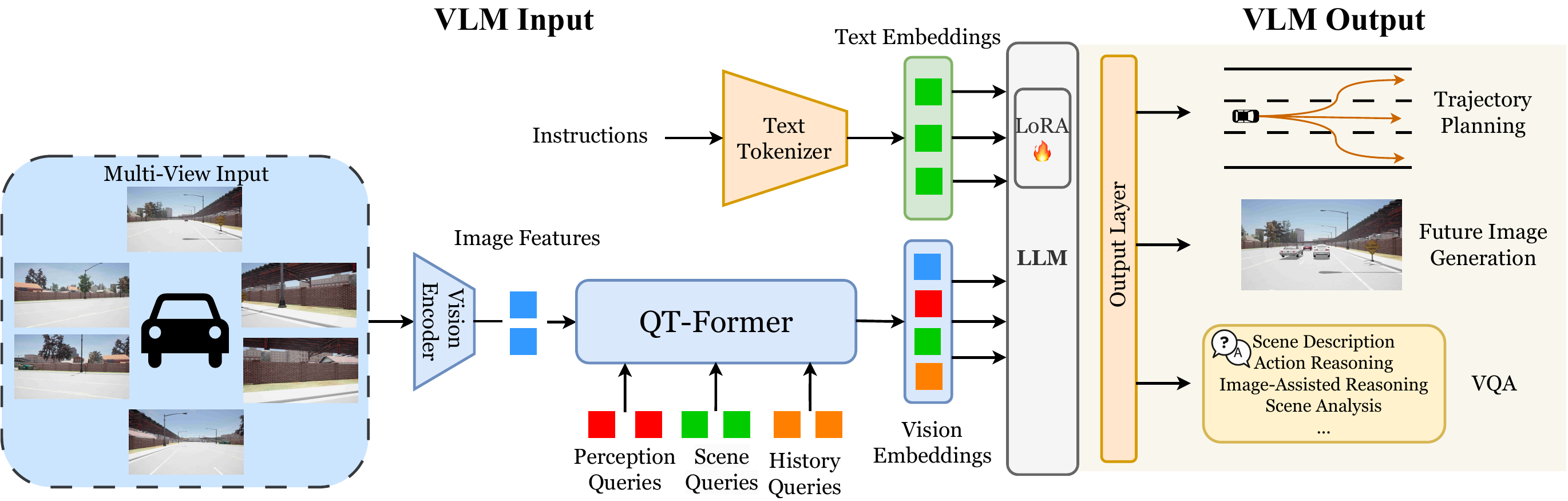}
    \caption{The pipeline of our UniDrive-WM framework. The pipeline consists of: (1) a QT-Former-based encoder to extract historical context and multi-view visual input; (2) the LLM for performing a reasoning task; and (3) the output layer, which generates the planning trajectory and future image prediction and bridges the gaps between the planning space, image space and reasoning space. For the output layer, we provide a detailed analysis in Fig.~\ref{fig:img_gen}.  } 
    
    \label{fig:pipeline}
\end{figure*}

In this section, we present our unified understanding, generation and planning framework for autonomous driving. The pipeline is shown in Fig.~\ref{fig:pipeline}.
We begin with the formulation of our method, followed by a detailed analysis of the system architecture. 
We then analyze the model architecture for planning and image generation. 
For image generation, we explore both (1) autoregressive image generation in a discrete visual representation and (2) diffusion-based generation in a continuous visual representation, and discuss their performance and impact on autonomous driving tasks. 
The task is defined as jointly predicting the future scene state and planning trajectory:
\begin{equation}
\{\hat{\mathbf{s}}_{t+n}, \hat{\mathbf{a}}_{t:t+m}\} 
\sim P_\theta \big( \mathbf{s}_{1:t},\, \mathbf{a}_{1:t-1},\, l \big),
\end{equation}
which can be further decomposed as:
\begin{equation}
\begin{gathered}
\left\{\hat{\mathbf{a}}_t, \ldots, \hat{\mathbf{a}}_{t+m}\right\}
\sim P_\theta\!\left(\left\{\mathbf{a}_t, \ldots, \mathbf{a}_{t+m}\right\} \mid \mathbf{s}_t, l\right), \\
\hat{\mathbf{s}}_{t+n}
\sim P_\theta\!\left(\mathbf{s}_{t+n} \mid \mathbf{s}_t, \hat{\mathbf{a}}_t, \ldots, \hat{\mathbf{a}}_{t+m}, l\right),
\end{gathered}
\end{equation}

\noindent
where $\mathbf{s}_t$ represents the multi-modal state information at time $t$, 
including multi-view images, historical context, and perception features; 
the model predicts the future front-view image $\hat{\mathbf{I}}_{t+n}$ 
as part of the future state $\hat{\mathbf{s}}_{t+n}$, 
and $\mathbf{a}_{t+m}$ denotes the predicted trajectory waypoint at time $t+m$; 
$l$ is the high-level language or instruction condition. 
The model therefore aims to jointly reason about the future world state and plan consistent trajectories under a unified vision–language world model.

\subsection{Vision Language Model}
A key component of a world model is understanding the current state and predicting the future state. Therefore, leveraging the Vision Language Model's reasoning and understanding ability, UniDrive-WM understands the current scene and instructs the trajectory planner and image generation module. We build our model upon ORION~\citep{fu2025orion}, a VLM-based model for the autonomous driving planning task.

\noindent \textbf{Vision Encoder} For the vision encoder, we adopt the QT-Former used by previous methods~\citep{fu2025orion}. Specifically, two sets of learnable queries are processed through self-attention (SA) to exchange information and interact with image features $F_m$ with 3D positional encoding in the cross-attention module. After that, the perception queries are fed into multiple auxiliary heads for object detection, including critical objects, lanes, traffic states and motion prediction of dynamic objects. Additionally, we use a set of history queries $Q_h \in \mathbb{R}^{(N_h \times C_q)}$, where $N_h$ denotes the number of history queries and $C_q$ represents the feature dimension, and a memory bank $M \in \mathbb{R}^{\left(N_h \times n\right) \times C_q}$ to retrieve and store historical information from the past $n$ frames. The QT-Former is formulated as:
\begin{equation}
Q_h = \mathrm{CA}\left(Q_h, M+P_t, M+P_t\right),\quad \hat{Q}_h = \mathrm{CA}\left(Q_h, Q_s, Q_s\right).
\end{equation}
where $\mathrm{CA}$ denotes the Cross-Attention operation, and $P_t$ denotes the relative timestamp embedding at the current timestep $t$. The updated history queries $\hat{Q}_h$ are stored in the memory bank $M$, formulated as:
\begin{equation}
M=\left[\hat{Q}_h^{t-n+1}, \cdots, \hat{Q}_h^{t-1}, \hat{Q}_h^t\right]
\end{equation}
Finally, a two-layer Multi-layer Perceptron (MLP) is used to convert the updated history queries $\hat{Q}_h$ and current scene features $Q_s$ to history tokens $x_h$ and scene tokens $x_s$ in the reasoning space of the LLM.

\noindent\textbf{Large Language Model.}
The large language model (LLM) serves as the reasoning core of our framework, enabling text-based understanding and high-level reasoning tasks such as scene description, visual question answering, and action reasoning. 
We fine-tune the LLM using LoRA~\citep{hu2022lora} to efficiently adapt it to domain-specific reasoning and instruction-following within autonomous driving scenarios.

\subsection{Trajectory Planner}

The trajectory planner establishes a differentiable connection between the semantic reasoning space of the VLM and the numerical action space of trajectory prediction through distribution learning in a latent space. 
We formulate trajectory generation as a conditional distribution problem, where the planner models a multi-modal distribution over future trajectories conditioned on high-level reasoning embeddings from the VLM:
\begin{equation}
p_\theta(\mathbf{a}_{t:t+m} \mid \mathbf{s}_t,\, \mathbf{h}_{\text{VLM}}),
\end{equation}
where $\mathbf{a}_{t:t+m}$ denotes the predicted trajectory waypoints, 
$\mathbf{s}_t$ represents the current state embedding, 
and $\mathbf{h}_{\text{VLM}}$ denotes the high-level reasoning embedding (hidden representation) extracted from the VLM, which encodes the fused multi-view observations, temporal history, and textual instructions. 
A latent variable $\mathbf{z}_a$ is introduced to capture the stochasticity of future motion in a Gaussian latent space, 
and the planner learns to decode diverse yet semantically consistent trajectories from $\mathbf{z}_a$. 
In practice, we omit the explicit KL regularization term used in conventional VAE objectives, 
as we empirically find it may destabilize training in large-scale multimodal setups. 
This design provides a stable and differentiable bridge between language-based reasoning and continuous trajectory prediction.

\begin{figure*}[t!]
    \centering
    \includegraphics[width=\linewidth]{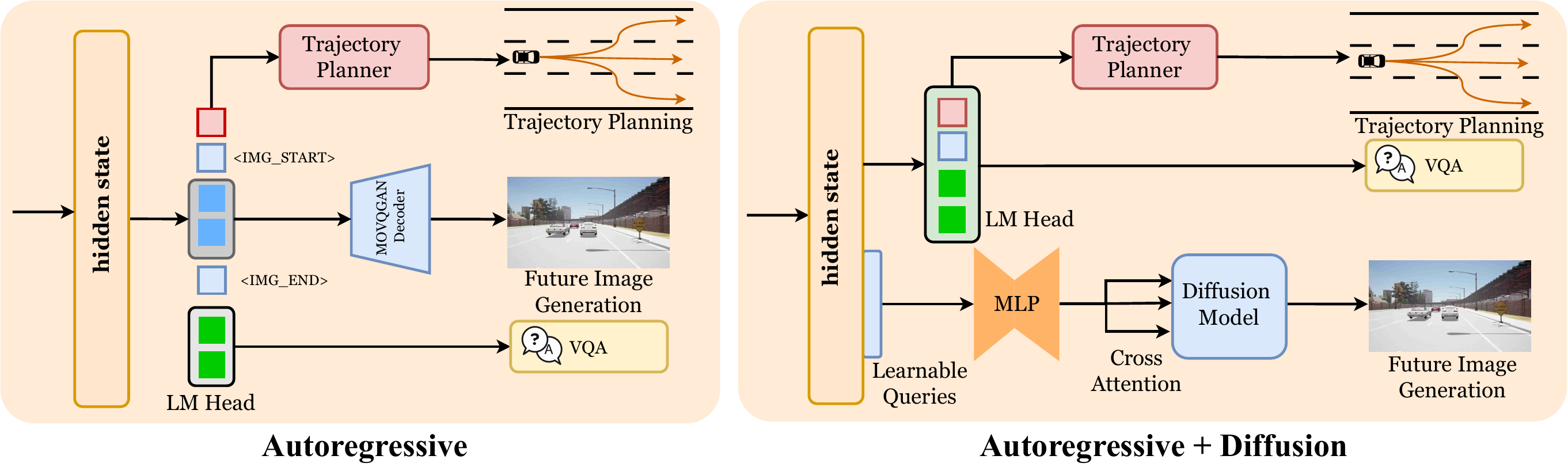}
    \caption{The two design choices for image generation in a unified multimodal model. For future image prediction, we use both Autoregressive and Autoregressive+Diffusion architectures. (a) Left: Autoregressive architecture; (b) Right: AR+Diffusion architecture.} 
    \label{fig:img_gen}
\end{figure*}

\subsection{Future Image Generation}

Since driving scenes are continuous in nature, autonomous driving datasets naturally contain dense future frames, allowing us to leverage abundant video data to improve image generation quality.
Inspired by recent advances in unifying image understanding and generation, we adopt two different image generation architectures for future image prediction, as illustrated in Fig.~\ref{fig:img_gen}. 
We discuss the design choices involved in building the understanding, planning, and image generation modules within a unified VLM framework, and analyze their impact on planning performance.

Formally, the future image prediction task can be formulated as:
\begin{equation}
p_\theta(\hat{\mathbf{I}}_{t+n} \mid \mathbf{s}_t,\, \mathbf{h}_{\text{VLM}}),
\end{equation}
where $\mathbf{s}_t$ denotes the multi-modal state information (including multi-view observations, temporal history, and perception features), 
and $\mathbf{h}_{\text{VLM}}$ is the high-level reasoning embedding extracted from the VLM that encodes both visual and textual context. 
The model aims to predict the future front-view frame $\hat{\mathbf{I}}_{t+n}$ conditioned on these multi-modal context embeddings.
This unified probabilistic formulation covers both the discrete \textbf{autoregressive (AR)} and the continuous \textbf{AR+Diffusion} architectures, 
which instantiate different parameterizations of the same conditional distribution $p_\theta(\cdot)$. 
We next describe the two paradigms instantiated under this formulation.

\subsubsection{Autoregressive Generation(Discrete Representation).} As shown in Fig.~\ref{fig:img_gen}(a), for discrete image generation, we first activate the VLM's visual generation ability by enlarging its codebook. 
Assume that the language–vision model uses a joint codebook $Z = Z_L + Z_I$. 
Through enlarging this codebook, the VLM can autoregressively perform multi-modal next-token prediction in both language and vision spaces. 
We introduce special tokens $\langle \mathit{image\_start} \rangle$ and $\langle \mathit{image\_end} \rangle$ to indicate the boundaries of visual token sequences and when to use the vision head.
Specifically, based on the input visual tokens and the planning token 
$\mathbf{a}_{t:t+m}$, the model autoregressively performs next-token prediction, represented as:
\begin{equation}
P_\theta(q_{1:H\!\cdot\!W} \mid \mathbf{s}_t,\, \mathbf{a}_{t:t+m})
= \prod_{i=1}^{H\!\cdot\!W}
P_\theta\!\left(q_i \mid q_{<i},\, \mathbf{s}_t,\, \mathbf{a}_{t:t+m}\right),
\end{equation}

\noindent
\textit{where} $q_i$ are the visual tokens, 
$\mathbf{s}_t$ represents the current multi-modal state (including visual observations, temporal history, and perception features), 
and $\mathbf{a}_{t:t+m}$ denotes the corresponding planning token that conditions the generation of subsequent visual tokens.

\subsubsection{AR+Diffusion Generation (Continuous Representation).}
Apart from fully autoregressive generation, we also design an \textbf{Autoregressive + Diffusion} architecture for improved visual quality, as shown in Fig.~\ref{fig:img_gen}(b).

The diffusion-based generation can be formulated as a conditional denoising process:
\begin{equation}
p_\theta(\hat{\mathbf{I}}_{t+n} \mid \mathbf{s}_t,\, \mathbf{h}_{\text{VLM}}),
\end{equation}
\noindent
where the decoder is conditioned on the high-level reasoning embedding $\mathbf{h}_{\text{VLM}}$ and the latent image features produced by the autoregressive module. 
Starting from Gaussian noise $\mathbf{X}_0 \!\sim\! \mathcal{N}(0,1)$, the diffusion process gradually refines $\mathbf{X}_0$ toward the latent embedding $\mathbf{X}_1$ of the ground-truth future image.

After extracting continuous visual embeddings, we employ an autoregressive transformer to generate corresponding latent image features conditioned on $\mathbf{h}_{\text{VLM}}$. 
Given an input instruction, the prompt is tokenized and mapped into a sequence of text embeddings $\mathbf{C} = [c_1, \ldots, c_n]$ before the LM head. 
To enable visual latent inference, we append a learnable query token $\mathbf{Q}$ to the sequence, where $\mathbf{Q}$ is randomly initialized and updated throughout training. 
The transformer processes the combined sequence $[\mathbf{C}; \mathbf{Q}]$, during which $\mathbf{Q}$ attends to the semantic context encoded in $\mathbf{h}_{\text{VLM}}$ and aggregates features relevant for image synthesis. 
The output query token, denoted as $\mathbf{Q}^*$, serves as the predicted visual latent representation and is supervised to match the ground-truth image embedding $\mathbf{X}$ extracted by the vision encoder. 
To align $\mathbf{Q}^*$ with $\mathbf{X}$, we use a flow-matching objective that models the continuous feature distribution. 
Given the ground-truth image feature $\mathbf{X}_1$ and the text-conditioned latent query $\mathbf{Q}$, 
we sample a timestep $t \sim \mathcal{U}(0,1)$ and a noise vector $\mathbf{X}_0 \sim \mathcal{N}(0,1)$. 
A latent point along the interpolation path is computed as
\begin{equation}
\mathbf{X}_t = t\mathbf{X}_1 + (1-t)\mathbf{X}_0 ,
\end{equation}
and the corresponding target velocity is
\begin{equation}
\mathbf{V}_t = \mathbf{X}_1 - \mathbf{X}_0 .
\end{equation}
The diffusion transformer predicts the velocity $\mathbf{V}_\theta(\mathbf{X}_t, \mathbf{Q}, t)$, and the flow-matching loss is
\begin{equation}
\mathcal{L}_{\text{FM}} = 
\mathbb{E}
\left[
\left\|
\mathbf{V}_\theta(\mathbf{X}_t, \mathbf{Q}, t) - \mathbf{V}_t
\right\|^2
\right].
\end{equation}

\noindent
Notably, we freeze the vision encoder weights and fine-tune only the diffusion decoder. 
Although the decoder adopts a diffusion architecture, it is trained with a deterministic reconstruction loss rather than probabilistic sampling objectives. 
Consequently, during inference, the model performs deterministic reconstruction, which reduces diversity but ensures stable and accurate prediction of future frames in autonomous driving scenarios. 
We additionally apply CLIP-based supervision between the decoded future image and the ground-truth image to maintain semantic alignment, represented as:
\begin{equation}
\mathcal{L}_{\text{CLIP}}
= 1 - \frac{\left\langle E_{\text{clip}}(I_{\rm pred}), E_{\text{clip}}(I_{\rm gt})\right\rangle}{\left\|E_{\text{clip}}(I_{\rm pred})\right\|_2\,\left\|E_{\text{clip}}(I_{\rm gt})\right\|_2},
\end{equation}
\noindent
where $\langle \cdot,\cdot \rangle$ denotes the Euclidean inner product, 
$E_{\text{clip}}(\cdot)$ denotes the embedding produced by the CLIP image encoder, 
$I_{\rm pred}$ is the predicted image, and $I_{\rm gt}$ is the ground-truth image. This diffusion pathway complements the discrete autoregressive generation branch by providing a continuous, geometry-aware latent space for future frame synthesis, leading to stable and semantically consistent visual predictions within the unified world-model framework.

\subsection{Joint Image Generation and Planning}

To achieve joint image generation and planning, we place the planning token immediately before the image tokens, so that the generation of each image token is conditioned on the generated planning representation. 
In this way, image generation is jointly conditioned on the current-state visual embeddings and the predicted trajectory of the future state, enabling the synthesis of consistent future frames aligned with planned motion.

For the autoregressive generation, image token prediction is treated analogously to language generation and supervised through teacher-forcing with a cross-entropy loss. 
We adopt the pretrained QT-Former from ORION~\citep{fu2025orion} and freeze its detection head during training. 
For the planning branch, following the implementations of VAD~\citep{jiang2023vad} and ORION~\citep{fu2025orion}, we train only the planning head with:
\begin{equation}
\mathcal{L}_{\text{plan}} = \mathcal{L}_{\text{col}} + \mathcal{L}_{\text{bd}} + \mathcal{L}_{\text{mse}},
\end{equation}
\noindent
where $\mathcal{L}_{\text{col}}$ denotes the collision loss, $\mathcal{L}_{\text{bd}}$ the boundary loss, and $\mathcal{L}_{\text{mse}}$ the mean squared error loss.

For the autoregressive architecture, the overall objective is represented as:
\begin{equation}
\mathcal{L} = \mathcal{L}_{\text{CE}} + \mathcal{L}_{\text{plan}},
\end{equation}
\noindent
while for the autoregressive + diffusion architecture, where a learnable latent query in the latent space serves as the conditional embedding for image generation, the full objective becomes:
\begin{equation}
\mathcal{L} = \mathcal{L}_{\text{CE}} + \mathcal{L}_{\text{plan}} + \mathcal{L}_{\text{FM}} + \mathcal{L}_{\text{CLIP}},
\end{equation}
\noindent
where $\mathcal{L}_{\text{FM}}$ is the flow-matching loss and $\mathcal{L}_{\text{CLIP}}$ is the CLIP-based semantic alignment loss.

\section{Experiments}
% 
% We introduce our method and  
\subsection{Training Details}
We follow ORION~\citep{fu2025orion} for the detection training setup and conduct all experiments on 8$\times$NVIDIA H200 GPUs. Following OmniDrive~\citep{wang2025omnidrive}, we use EVA-02-L~\citep{fang2024eva} as the vision encoder and adopt Vicuna~1.5~\citep{zheng2023judging} as the base LLM, fine-tuned with LoRA (rank=16, $\alpha$=16). The numbers of scene, perception, and history queries are set to 512, 600, and 16, respectively, with a memory bank size of 16 frames.

Input images are augmented and resized to $640\times640$. For future frame prediction, we use a $192\times128$ resolution for the autoregressive branch and $512 \times 1024$ for the AR+Diffusion branch to balance quality and speed. The diffusion decoder uses $64$ latent learnable query tokens. Additional hyperparameters and implementation details are provided in the Appendix.

\subsection{Dataset}
We train and evaluate UniDrive-WM on the Bench2Drive~\citep{jia2024bench2drive} dataset and the nuScenes~\citep{caesar2020nuscenes} dataset. For Bench2Drive, following prior work, we use the base split of 1000 driving scenes: 950 for training and 50 for open-loop validation. Each scene covers roughly 150\,m of continuous driving in a distinct traffic scenario. For closed-loop evaluation, we follow the official protocol consisting of 220 short routes spanning 44 interactive scenarios (five routes per scenario). For nuScenes, we adopt the official train/validation split for all experiments.

% Bench2Drive is a closed-loop benchmark built on CARLA v2~\citep{dosovitskiy2017carla}.

\subsection{Training Pipeline}
To enable planning and image generation while preserving the VQA capability of the underlying VLM, we adopt a two-stage training strategy.

\noindent\textbf{Stage 1: Joint Planning and Image Generation}
We train the full model end-to-end, with the LLM updated via LoRA. 
For the autoregressive generator, the image-start token is placed immediately after the planning token to enforce planning-conditioned autoregressive decoding. 
For the AR+Diffusion architecture, the 64 learnable latent query tokens are appended after the planning features in the latent space to condition the diffusion decoder.

\noindent\textbf{Stage 2: Joint Planning, Image Generation, and VQA}
We continue training with mixed VQA and driving data while keeping the same architectural setup as Stage 1. This stage reinforces the alignment of the vision–language–planning space by jointly optimizing VQA, trajectory planning, and future image prediction, enabling unified multimodal reasoning within a single framework.

\begin{table}[t]
\centering
\caption{Closed-loop and Open-loop Results of E2E-AD Methods on the Bench2Drive base set. C/L refers to camera/LiDAR. Avg. L2 is averaged over a 2-second prediction horizon sampled at 2 Hz, similar to UniAD. * denotes expert feature distillation. NC: navigation command, TP: target point, DS: Driving Score, SR: Success Rate.}
\scriptsize
\label{tab:close-loop}
\resizebox{\linewidth}{!}{%
\begin{tabular}{lccccccc}
\toprule
\multirow{2}{*}{Method} & \multirow{2}{*}{Condition} & \multirow{2}{*}{Modality} & \multicolumn{4}{c}{Closed-loop Metric} & Open-loop Metric \\
\cmidrule(lr){4-7} \cmidrule(lr){8-8}
 &  &  & DS$\uparrow$ & SR(\%)$\uparrow$ & Efficiency$\uparrow$ & Comfortness$\uparrow$ & Avg. L2 $\downarrow$ \\
\midrule
TCP*~\citep{wu2022trajectory} & TP & C & 40.70 & 15.00 & 54.26 & 47.80 & 1.70 \\
TCP-ctrl* & TP & C & 30.47 & 7.27 & 55.97 & 51.51 & - \\
TCP-traj* & TP & C & 59.90 & 30.00 & 76.54 & 18.08 & 1.70 \\
TCP-traj w/o distillation & TP & C & 49.30 & 20.45 & 78.78 & 22.96 & 1.96 \\
ThinkTwice*~\citep{Jia_2023_CVPR} & TP & C & 62.44 & 31.23 & 69.33 & 16.22 & 0.95 \\
DriveAdapter*~\citep{jia2023driveadapter} & TP & C\&L & 64.22 & 33.08 & 70.22 & 16.01 & 1.01 \\
\midrule
AD-MLP~\citep{zhai2023rethinking} & NC & C & 18.05 & 0.00 & 48.45 & 22.63 & 3.64 \\
UniAD-Tiny~\citep{hu2023_uniad} & NC & C & 40.73 & 13.18 & 123.92 & 47.04 & 0.80 \\
UniAD-Base~\citep{hu2023_uniad} & NC & C & 45.81 & 16.36 & 129.21 & 43.58 & 0.73 \\
VAD~\citep{jiang2023vad} & NC & C & 42.35 & 15.00 & 157.94 & 46.01 & 0.91 \\
MomAD~\citep{song2025don} & NC & C & 44.54 & 16.71 & 170.21 & 48.63 & 0.87 \\
GenAD~\citep{zheng2024genad} & NC & C & 44.81 & 15.90 & - & - & - \\
DriveTransformer-Large~\citep{jia2025drivetransformer} & NC & C & 63.46 & 35.01 & 100.64 & 20.78 & 0.62 \\
ORION~\citep{fu2025orion} & NC & C & 77.74 & 54.62 & 151.48 & 17.38 & 0.68 \\
\midrule
\textbf{Ours(AR)} & NC & C & \textbf{79.22} & \textbf{56.36} & 158.44 & 28.01 & 0.64 \\
\textbf{Ours(AR+Diffusion)} & NC & C & \textbf{79.31} & \textbf{56.42} & 158.65 & 27.93 & 0.63 \\
\bottomrule
\end{tabular}

}
\end{table}

\begin{table*}[t]
\centering
\scriptsize

\caption{Open-Loop Planning and perception metrics on Bench2Drive. Lower is better for planning errors; higher is better for mAP/NDS.}
\label{tab:open-loop}

\resizebox{\textwidth}{!}{
\begin{tabular}{lccc|ccc|cccccc}
\toprule
\multirow{2}{*}{Method} 
& \multicolumn{3}{c|}{L2 (m)$\downarrow$} 
& \multicolumn{3}{c|}{Box + Collision(\%)$\downarrow$} 
& mAP$\uparrow$ 
& mATE$\downarrow$ & mASE$\downarrow$ & mAOE$\downarrow$ & mAVE$\downarrow$ & NDS$\uparrow$ \\
\cmidrule(lr){2-4} \cmidrule(lr){5-7}
& 1s & 2s & 3s 
& 1s & 2s & 3s 
&  &  &  &  &  &  \\
\midrule
BEVFormer~\citep{li2022bevformer} & -- & -- & -- & -- & -- & -- 
& 0.616 & 0.372 & 0.079 & 0.044 & 0.808 & 0.642 \\

UniAD~\citep{hu2023_uniad} & 0.521 & 1.265 & 2.140 
& 0.770 & 3.87 & 7.91 
& 0.121 & 0.518 & 0.170 & 0.096 & 0.977 & 0.316 \\

VAD~\citep{jiang2023vad} & 0.454 & 0.912 & 1.477 
& 0.102 & 0.202 &  0.296
& 0.509 & 0.385 & 0.0854 & 0.0308 & 0.594 & 0.604 \\

Orion~\citep{fu2025orion}
& 0.268 & 0.631 & 1.129 
& 0.213 & 0.467 & 0.743 
& 0.646 & 0.366 & 0.0765 & 0.0271 & 0.258 & 0.723 \\

\textbf{Ours(AR)} 
& \textbf{0.247} & \textbf{0.598} & \textbf{1.079}
& \textbf{0.198} & \textbf{0.435} & \textbf{0.668}
& \textbf{0.663} & \textbf{0.335} & \textbf{0.0740} & \textbf{0.0262} & \textbf{0.249} & \textbf{0.746} \\

\textbf{Ours(AR+Diff)} 
& \textbf{0.241} & \textbf{0.589} & \textbf{1.066} 
& \textbf{0.192} & \textbf{0.426} & \textbf{0.657} 
& \textbf{0.675} & \textbf{0.328} & \textbf{0.0733} & \textbf{0.0254} & \textbf{0.243} & \textbf{0.755} \\

\bottomrule
\end{tabular}}
\end{table*}

\begin{table}[t]
\centering
\caption{Further Open-Loop evaluation on nuScenes.}
\label{tab:nuscenes}
\scriptsize
\resizebox{\textwidth}{!}{%
\begin{tabular}{lcccccccc}
\toprule
\textbf{Metric}
& VAD-Base~\citep{jiang2023vad}
& Doe-1~\citep{doe}
& Drive-VLM~\citep{tian2024drivevlm}
& OmniDrive~\citep{wang2025omnidrive}
& ORION~\citep{fu2025orion}
& FSDrive~\citep{zeng2025futuresightdrive}
& \textbf{Ours(AR)}
& \textbf{Ours(AR+Diff)}  \\
\midrule
Avg.\ L2 (m)$\downarrow$
& 1.25
& 0.70
& 0.40
& 0.84
& 0.34
& 0.60
& \textbf{0.30} 
& \textbf{0.29} \\
Avg.\ col (\%)$\downarrow$
& 1.09
& 0.21
& 0.27
& 0.94
& 0.37
& \textbf{0.19}
& 0.31 
& 0.31 \\
\bottomrule
\end{tabular}%
}
\end{table}

\subsection{Results}
\subsubsection{Evaluation of Trajectory Planning Results}

We evaluate our results on the base validation set of Bench2Drive and nuScenes. We report both open-loop and closed-loop evaluation results. As shown in Tab.~\ref{tab:close-loop}, for closed-loop evaluation results on Bench2Drive, our method outperforms all the other methods that rely on target point and navigation command as conditions. Results show that our method achieves better performance compared with previous end-to-end methods and VLM-guided planning methods, which demonstrates the effectiveness of the image generation modality in boosting the performance of the planning task.

We also evaluate the open-loop performance on Bench2Drive in Tab.~\ref{tab:open-loop} and nuScenes in Tab.~\ref{tab:nuscenes}. Notably, UniAD~\citep{hu2023_uniad} computes L2 metrics and collision rate at each timestep, whereas VAD~\citep{jiang2023vad} and ORION~\citep{fu2025orion} consider the average of all previous time steps. We follow the evaluation method of VAD and ORION to evaluate our performance. Compared with previous methods that perform either VLM-guided planning or end-to-end planning, our unified method achieves better performance on both the planning and detection tasks, which shows the effectiveness of future image prediction in improving the performance of the planning task. These consistent gains across both Bench2Drive (synthetic) and NuScenes (real-world) further indicate the strong generalization ability of our method.

\begin{table}[t!]
\centering
\footnotesize
\caption{Future frame generation results on the Bench2Drive~\citep{jia2024bench2drive} and nuScenes~\citep{caesar2020nuscenes} datasets. AR: Autoregressive. AR+Diff: Autoregressive+Diffusion. Lower FID indicates better visual fidelity.}
\label{tab:future_frames}

\resizebox{\linewidth}{!}{%
\setlength{\heavyrulewidth}{1.5pt}
\begin{tabular}{lccccccccc}
\toprule
\textbf{Method} & \textbf{Dataset} & {\begin{tabular}{c} DriveGAN~\citep{kim2021drivegan} \\ {[CVPR21]} \end{tabular}} & {\begin{tabular}{c} DriveDreamer~\citep{wang2023drivedreamer} \\ {[ECCV24]} \end{tabular}} & {\begin{tabular}{c} Drive-WM~\cite{wang2023driving} \\ {[CVPR24]} \end{tabular}} & {\begin{tabular}{c} GEM~\cite{hassan2025gem} \\ {[CVPR25]} \end{tabular}} & {\begin{tabular}{c} Doe-1~\cite{doe} \\ {[arXiv24]} \end{tabular}} & {\begin{tabular}{c} FSDrive~\citep{zeng2025futuresightdrive} \\ {[NeurIPS25]} \end{tabular}} & \textbf{Ours(AR)} & \textbf{Ours (AR+Diff)} \\ 
\midrule
\textbf{Type} & -- & GAN & Diffusion & Diffusion & Diffusion & AR & AR & AR & AR+Diffusion \\
\midrule
\multirow{2}{*}{\textbf{FID} $\downarrow$} & Bench2Drive & 62.3 & 42.8 & 17.8 & 13.2 & 18.6 & 9.3 & \textbf{7.2} & \textbf{6.6} \\
 & nuScenes & 73.4 & 52.6 & 15.8 & 10.5 & 15.9 & 10.1 & \textbf{7.8} & \textbf{7.3} \\
\bottomrule
\end{tabular}
}
\end{table}

\subsubsection{Evaluation of Image Generation Results}

\begin{figure}[t!]
    \centering
    \includegraphics[width=\linewidth]{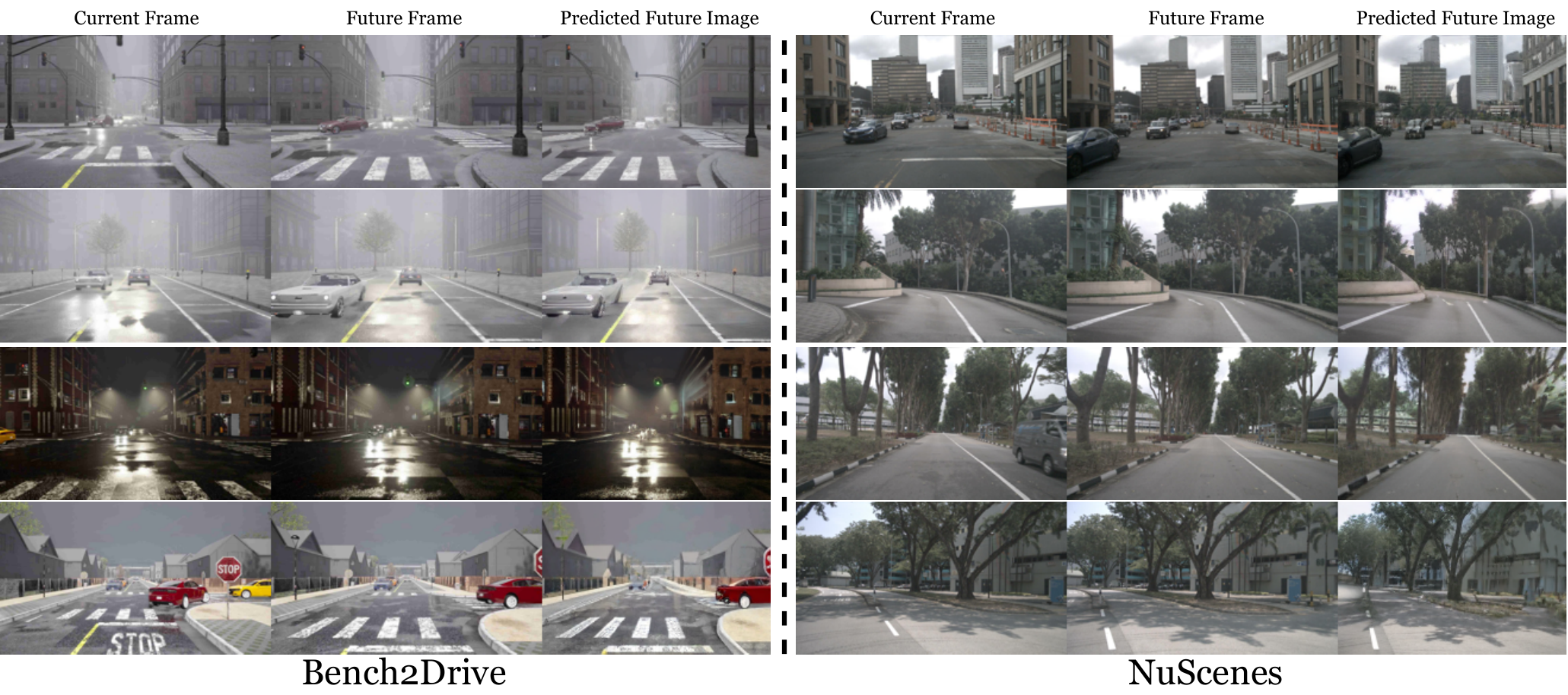}
    \caption{Qualitative future image prediction results on Bench2Drive and nuScenes using the autoregressive (AR) architecture.}
    
    \label{fig:gen_ar}
\end{figure}

\begin{figure}[ht!]
    \centering
    \includegraphics[width=\linewidth]{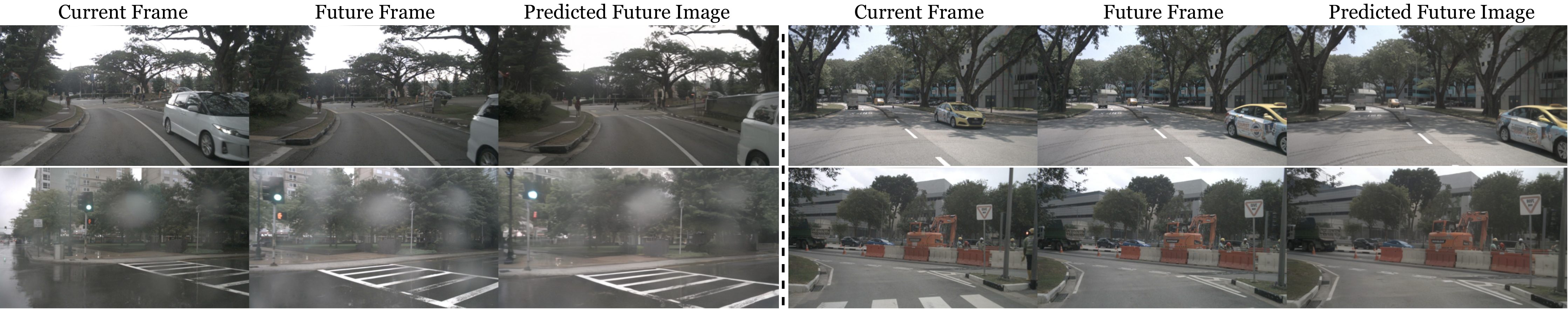}
    \caption{Qualitative future image prediction results on nuScenes using the AR+Diffusion architecture.} 
    \label{fig:gen_ardiff}
\end{figure}

We evaluate UniDrive-WM on both the autoregressive and AR+Diffusion image generation branches. Qualitative results are shown in Fig.~\ref{fig:gen_ar} and Fig.~\ref{fig:gen_ardiff}, and quantitative results are provided in Tab.~\ref{tab:future_frames}. Across both settings, our method produces visually coherent future frames that closely match the ground-truth scene evolution and remain consistent with the predicted trajectory. These properties translate into competitive or superior image generation quality, as reflected by low FID scores.

A particularly relevant comparison is FSDrive~\citep{zeng2025futuresightdrive}, a recent unified vision--planning framework that conditions planning on image features. Unlike FSDrive, our approach conditions image generation directly on the planning tokens, enabling planning-conditioned future frame prediction in both the AR and AR+Diffusion branches. This tighter coupling between planning and visual prediction improves trajectory alignment and yields future frames that more accurately follow the intended motion, contributing to our improved FID performance.

\begin{table}[t]
\centering
\caption{Results on the DriveLM~\citep{sima2024drivelm} GVQA benchmark.}
\label{tab:drivelm_gvqa}
\small

\resizebox{\textwidth}{!}{%
\begin{tabular}{l|ccccccc}
\toprule
\textbf{Method} 
& \textbf{Accuracy $\uparrow$} 
& \textbf{ChatGPT $\uparrow$} 
& \textbf{BLEU\_1 $\uparrow$} 
& \textbf{ROUGE\_L $\uparrow$} 
& \textbf{CIDEr $\uparrow$} 
& \textbf{Match $\uparrow$} 
& \textbf{Final Score $\uparrow$} \\
\midrule
DriveLM baseline~\cite{sima2024drivelm} & 0.00 & 0.65 & 0.05 & 0.08 & 0.10 & 0.28 & 0.32 \\
Cube-LLM~\cite{cho2024language}         & 0.39 & \textbf{0.89} & 0.16 & 0.20 & \textbf{0.31} & 0.39 & 0.50 \\
TrackingMeetsLMM~\cite{ishaq2025tracking} & 0.60 & 0.58 & 0.72 & 0.72 & 0.04 & 0.36 & 0.52 \\
SimpleLLM4AD~\cite{zheng2024simplellm4ad} & 0.66 & 0.57 & 0.76 & 0.73 & 0.15 & 0.35 & 0.53 \\
OmniDrive~\cite{wang2025omnidrive}      & 0.70 & 0.65 & 0.52 & 0.73 & 0.13 & 0.37 & 0.56 \\
FSDrive~\citep{zeng2025futuresightdrive} & 0.72 & 0.63 & 0.76 & 0.74 & 0.17 & 0.39 & 0.57 \\
\midrule
\textbf{UniDrive-WM(ours)} & \textbf{0.73} & 0.67 & \textbf{0.77} & \textbf{0.76} & 0.20 & \textbf{0.40} & \textbf{0.59} \\
\bottomrule
\end{tabular}
}
\end{table}

\subsubsection{Evaluation on VQA}

Following prior work, we report our Visual Question Answering (VQA) results in Tab.~\ref{tab:drivelm_gvqa}, using DriveLM's GVQA~\citep{sima2024drivelm} dataset. Our model achieves competitive performance against prior state-of-the-art methods on the understanding task, demonstrating its effectiveness in driving-domain-specific understanding.

\subsection{Comparison of AR and AR+Diffusion Architecture}

\noindent\textbf{Speed} The AR branch is intentionally lightweight (discrete tokens, shallow decoder); the AR+Diff branch is heavier (diffusion decoder) and mainly intended for high-fidelity generation. In open-loop evaluation on Bench2Drive, the inference speed of AR is 2 fps and AR+Diff is 0.4 fps on an A100. We regard speeding up the AR+Diff inference as future work.

\noindent\textbf{Resolution and Compression Trade-offs} 
In the AR architecture, spatial resolution is tightly coupled to sequence length. Generating an $H \times W$ image with a downsampling factor $f$ requires $\frac{H}{f} \times \frac{W}{f}$ discrete tokens. Scaling up resolution quadratically increases the token count, which not only hits the LLM's $O(N^2)$ context limit but also severely slows down training convergence. In contrast, our AR+Diffusion branch decouples resolution from token length. The VLM only needs to predict a compact sequence of continuous latents, which the diffusion decoder maps to high-resolution outputs (e.g., $512 \times 1024$). This enables high-fidelity geometric forecasting without overwhelming the reasoning context window or hindering training efficiency.

\noindent\textbf{Image Generation and Complementary Roles}
Beyond resolution constraints, the two paradigms exhibit distinct characteristics in representation learning (Tab.~\ref{tab:future_frames}). The discrete quantization in the AR branch acts as an information bottleneck that provides strong semantic regularization. This makes its lightweight decoder highly robust for real-time, latency-sensitive closed-loop control. In contrast, the continuous pathway of the AR+Diffusion branch avoids token-by-token exposure bias and inherently preserves high-frequency geometric details. This yields higher-fidelity future frames and more precise long-term trajectory alignment (e.g., lower L2 errors and higher NDS). Implemented as parallel decoding heads on top of the same VLM planning backbone, the two branches offer a natural deployment split: AR is suited to fast, reactive planning where semantic stability is critical, while AR+Diffusion is preferable for offline evaluation or high-fidelity geometric forecasting in complex scenes. Overall, the two heads are complementary rather than competing, and together broaden the applicability of UniDrive-WM.

\begin{table}[t!]
\caption{Ablation study of the detection head, planning head, and image generation module for our AR architecture on Bench2Drive~\cite{jia2024bench2drive}.}
\label{tab:ablation_study}
\resizebox{\textwidth}{!}{
\setlength{\heavyrulewidth}{1.5pt} 
\begin{tabular}{ccc|ccc|ccc|cccccc}
\toprule
\multirow{2}{*}{Detection head} 
& \multirow{2}{*}{Planning head} 
& \multirow{2}{*}{Image Gen}
& \multicolumn{3}{c|}{L2 (m)$\downarrow$} 
& \multicolumn{3}{c|}{Box + Collision (\%)$\downarrow$} 
& mAP$\uparrow$ 
& mATE$\downarrow$ & mASE$\downarrow$ & mAOE$\downarrow$ & mAVE$\downarrow$ & NDS$\uparrow$ \\
\cmidrule(lr){4-6} \cmidrule(lr){7-9}
& & 
& 1s & 2s & 3s 
& 1s & 2s & 3s 
&  &  &  &  &  &  \\
\midrule
 & \checkmark &  \checkmark
& 0.482 & 1.135 & 2.146 
& 0.387 & 0.898 & 1.77 
& -- & -- & -- & -- & -- & -- \\
\checkmark & & \checkmark 
& -- & -- & -- 
& -- & -- & -- 
& 0.638 & 0.336 & 0.0726 & 0.0289 & 0.283 & 0.718 \\
\checkmark & \checkmark &  
& 0.269 & 0.632 & 1.130 
& 0.214 & 0.469 & 0.746 
& 0.642 & 0.367 & 0.0767 & 0.0274 & 0.258 & 0.721 \\
\checkmark & \checkmark & \checkmark 
& \textbf{0.247} & \textbf{0.598} & \textbf{1.079}
& \textbf{0.198} & \textbf{0.435} & \textbf{0.668}
& \textbf{0.663} & \textbf{0.335} & \textbf{0.0740} & \textbf{0.0262} & \textbf{0.249} & \textbf{0.746} \\
\bottomrule
\end{tabular}}
\end{table}

\begin{table}[t!]
\centering
\scriptsize
\caption{Ablation study on open-loop evaluation on Bench2Drive. For the AR architecture, we analyze the effect of swapping the order of the planning and generation heads.}

\label{tab:openloop-ablation-turn}
\begin{tabular}{lcc|c}
\toprule
Future Image Prediction
& Avg.\ L2 (m)$\downarrow$
& Avg.\ col(\%)$\downarrow$
& FID$\downarrow$ \\
\midrule
Generation + Planning & 0.67 & 0.45 & 9.1 \\
Planning + Generation (\textbf{Ours})
& \textbf{0.64} & \textbf{0.43} & \textbf{7.2} \\
\bottomrule
\end{tabular}
\end{table}

\subsection{Ablation Study}

\begin{wraptable}[7]{r}[-2pt]{0.5\textwidth}
\vspace{-25pt}
\raggedleft

\begin{minipage}[t]{0.98\linewidth}
\scriptsize
\caption{Additional ablation of VQA score on Chat-B2D.}
\label{tab:vqa}
\begin{tabular}{cccc}
\toprule
Image Generation & CIDEr $\uparrow$ & BLEU $\uparrow$ & ROUGE-L$\uparrow$ \\
\midrule
- & 65.7 & 52.4 & 77.5 \\
\checkmark & \textbf{66.7} & \textbf{53.5} & \textbf{78.6} \\
\bottomrule
\end{tabular}
\end{minipage}
\end{wraptable}

We ablate the detection supervision and the image generation module to assess their contributions to planning, shown in Tab.~\ref{tab:ablation_study}. Removing the image generation head noticeably degrades trajectory accuracy and increases collision cases, indicating that future frame prediction provides useful auxiliary signals for planning. Disabling detection supervision further harms planning performance by weakening perception quality. Together, these findings show that both perception and future frame prediction play complementary roles, and that jointly optimizing all components within a unified world model yields more reliable driving behavior.

We show experiments regarding changing the order of the planning and image generation modules in the AR architecture in Tab.~\ref{tab:openloop-ablation-turn}. \textbf{Planning $\rightarrow$ Generation (Ours)} shows improvements over \textbf{Generation $\rightarrow$ Planning} on planning-related metrics and a much larger gain in visual quality, suggesting that the prediction order matters; conditioning future prediction on planned trajectories aligns with the action$\rightarrow$observation causal direction and provides a natural mechanism to produce plan-consistent futures that better support planning.

We evaluate VQA performance on Bench2Drive~\cite{jia2024bench2drive} using the Chat-B2D QA annotations~\citep{fu2025orion}, shown in Tab.~\ref{tab:vqa}. Compared with the model variant without the image-generation module, incorporating image generation consistently improves all VQA metrics. This indicates that the future frame prediction provides additional structural cues that benefit question answering. Intuitively, better modeling of future visual states enhances the model’s understanding of the current scene, which aligns with human perception. We also show the visualization of the VQA result in open-loop evaluation in the supplementary material. These results demonstrate that our model not only comprehends the current scene, but also performs effective future-state reasoning, enabling it to produce reliable, decision-oriented answers even in complex driving scenarios and challenging weather conditions.

\section{Conclusion and Future Work}
We presented UniDrive-WM, a unified framework that integrates scene understanding, trajectory planning, and visual generation into a single world-model–driven pipeline. Leveraging a vision–language model (VLM) as the backbone, our method predicts future image frames from current and historical multi-view observations together with the planning tokens, and the planning-conditioned visual predictions in turn enhance trajectory planning by providing a visual forecast of the expected future scene. Experimental results demonstrate that our approach not only achieves high-fidelity, planning-conditioned image generation, but also yields significant gains in planning accuracy and collision reduction. By conditioning the VLM on rich multi-view inputs, temporal history, and perception features, UniDrive-WM establishes a coherent bridge between reasoning, action, and visual imagination. Looking ahead, we plan to extend the framework to more interactive and long-horizon driving scenarios, paving the way toward next-generation world models for autonomous driving.

% \clearpage\mbox{}Page \thepage\ of the manuscript.
% \clearpage\mbox{}Page \thepage\ of the manuscript.
% \clearpage\mbox{}Page \thepage\ of the manuscript.
% \clearpage\mbox{}Page \thepage\ of the manuscript.
% \clearpage\mbox{}Page \thepage\ of the manuscript. This is the last page.
\par\vfill\par

\bibliographystyle{splncs04}
\bibliography{main}

@String(CVPR  = {IEEE Conf. Comput. Vis. Pattern Recog.})

@String(ICCV  = {Int. Conf. Comput. Vis.})

@String(ICLR  = {Int. Conf. Learn. Represent.})

@String(CVPR  = {CVPR})

@String(ICCV  = {ICCV})

@String(ICLR  = {ICLR})

@article{jia2024bench2drive,
  title={Bench2drive: Towards multi-ability benchmarking of closed-loop end-to-end autonomous driving},
  author={Jia, Xiaosong and Yang, Zhenjie and Li, Qifeng and Zhang, Zhiyuan and Yan, Junchi},
  journal={Advances in Neural Information Processing Systems},
  volume={37},
  pages={819--844},
  year={2024}
}

@article{fu2025orion,
  title={Orion: A holistic end-to-end autonomous driving framework by vision-language instructed action generation},
  author={Fu, Haoyu and Zhang, Diankun and Zhao, Zongchuang and Cui, Jianfeng and Liang, Dingkang and Zhang, Chong and Zhang, Dingyuan and Xie, Hongwei and Wang, Bing and Bai, Xiang},
  journal={arXiv preprint arXiv:2503.19755},
  year={2025}
}

@article{team2024chameleon,
  title={Chameleon: Mixed-modal early-fusion foundation models},
  author={Team, Chameleon},
  journal={arXiv preprint arXiv:2405.09818},
  year={2024}
}

@article{xie2024showo,
  title={Show-o: One Single Transformer to Unify Multimodal Understanding and Generation},
  author={Xie, Jinheng and Mao, Weijia and Bai, Zechen and Zhang, David Junhao and Wang, Weihao and Lin, Kevin Qinghong and Gu, Yuchao and Chen, Zhijie and Yang, Zhenheng and Shou, Mike Zheng},
  journal={arXiv preprint arXiv:2408.12528},
  year={2024}
}

@inproceedings{wu2025janus,
  title={Janus: Decoupling visual encoding for unified multimodal understanding and generation},
  author={Wu, Chengyue and Chen, Xiaokang and Wu, Zhiyu and Ma, Yiyang and Liu, Xingchao and Pan, Zizheng and Liu, Wen and Xie, Zhenda and Yu, Xingkai and Ruan, Chong and others},
  booktitle={Proceedings of the Computer Vision and Pattern Recognition Conference},
  pages={12966--12977},
  year={2025}
}

@article{tong2024metamorph,
  title={Metamorph: Multimodal understanding and generation via instruction tuning},
  author={Tong, Shengbang and Fan, David and Zhu, Jiachen and Xiong, Yunyang and Chen, Xinlei and Sinha, Koustuv and Rabbat, Michael and LeCun, Yann and Xie, Saining and Liu, Zhuang},
  journal={arXiv preprint arXiv:2412.14164},
  year={2024}
}

@misc{chen2025blip3ofamilyfullyopen,
title={BLIP3-o: A Family of Fully Open Unified Multimodal Models-Architecture, Training and Dataset},
author={Jiuhai Chen and Zhiyang Xu and Xichen Pan and Yushi Hu and Can Qin and Tom Goldstein and Lifu Huang and Tianyi Zhou and Saining Xie and Silvio Savarese and Le Xue and Caiming Xiong and Ran Xu},
year={2025},
eprint={2505.09568},
archivePrefix={arXiv},
primaryClass={cs.CV},
url={https://arxiv.org/abs/2505.09568},
}

@article{chen2025blip3o,
  title={Blip3o-next: Next frontier of native image generation},
  author={Chen, Jiuhai and Xue, Le and Xu, Zhiyang and Pan, Xichen and Yang, Shusheng and Qin, Can and Yan, An and Zhou, Honglu and Chen, Zeyuan and Huang, Lifu and others},
  journal={arXiv preprint arXiv:2510.15857},
  year={2025}
}

@article{zhou2024transfusion,
  title={Transfusion: Predict the next token and diffuse images with one multi-modal model},
  author={Zhou, Chunting and Yu, Lili and Babu, Arun and Tirumala, Kushal and Yasunaga, Michihiro and Shamis, Leonid and Kahn, Jacob and Ma, Xuezhe and Zettlemoyer, Luke and Levy, Omer},
  journal={arXiv preprint arXiv:2408.11039},
  year={2024}
}

@article{zheng2023judging,
  title={Judging llm-as-a-judge with mt-bench and chatbot arena},
  author={Zheng, Lianmin and Chiang, Wei-Lin and Sheng, Ying and Zhuang, Siyuan and Wu, Zhanghao and Zhuang, Yonghao and Lin, Zi and Li, Zhuohan and Li, Dacheng and Xing, Eric and others},
  journal={Advances in neural information processing systems},
  volume={36},
  pages={46595--46623},
  year={2023}
}

@article{hu2022lora,
  title={Lora: Low-rank adaptation of large language models.},
  author={Hu, Edward J and Shen, Yelong and Wallis, Phillip and Allen-Zhu, Zeyuan and Li, Yuanzhi and Wang, Shean and Wang, Lu and Chen, Weizhu and others},
  journal={ICLR},
  volume={1},
  number={2},
  pages={3},
  year={2022}
}

@article{zhang2025epona,
  title={Epona: Autoregressive Diffusion World Model for Autonomous Driving},
  author={Zhang, Kaiwen and Tang, Zhenyu and Hu, Xiaotao and Pan, Xingang and Guo, Xiaoyang and Liu, Yuan and Huang, Jingwei and Yuan, Li and Zhang, Qian and Long, Xiao-Xiao and others},
  journal={arXiv preprint arXiv:2506.24113},
  year={2025}
}

@inproceedings{hassan2025gem,
  title={Gem: A generalizable ego-vision multimodal world model for fine-grained ego-motion, object dynamics, and scene composition control},
  author={Hassan, Mariam and Stapf, Sebastian and Rahimi, Ahmad and Rezende, Pedro and Haghighi, Yasaman and Br{\"u}ggemann, David and Katircioglu, Isinsu and Zhang, Lin and Chen, Xiaoran and Saha, Suman and others},
  booktitle={Proceedings of the Computer Vision and Pattern Recognition Conference},
  pages={22404--22415},
  year={2025}
}

@article{gao2024vista,
  title={Vista: A generalizable driving world model with high fidelity and versatile controllability},
  author={Gao, Shenyuan and Yang, Jiazhi and Chen, Li and Chitta, Kashyap and Qiu, Yihang and Geiger, Andreas and Zhang, Jun and Li, Hongyang},
  journal={Advances in Neural Information Processing Systems},
  volume={37},
  pages={91560--91596},
  year={2024}
}

@article{wei2024occllama,
  title={Occllama: An occupancy-language-action generative world model for autonomous driving},
  author={Wei, Julong and Yuan, Shanshuai and Li, Pengfei and Hu, Qingda and Gan, Zhongxue and Ding, Wenchao},
  journal={arXiv preprint arXiv:2409.03272},
  year={2024}
}

@article{wang2024occsora,
  title={Occsora: 4d occupancy generation models as world simulators for autonomous driving},
  author={Wang, Lening and Zheng, Wenzhao and Ren, Yilong and Jiang, Han and Cui, Zhiyong and Yu, Haiyang and Lu, Jiwen},
  journal={arXiv preprint arXiv:2405.20337},
  year={2024}
}

@article{zhang2023copilot4d,
  title={Copilot4d: Learning unsupervised world models for autonomous driving via discrete diffusion},
  author={Zhang, Lunjun and Xiong, Yuwen and Yang, Ze and Casas, Sergio and Hu, Rui and Urtasun, Raquel},
  journal={arXiv preprint arXiv:2311.01017},
  year={2023}
}

@article{zhang2024bevworld,
  title={BEVWorld: A Multimodal World Simulator for Autonomous Driving via Scene-Level BEV Latents},
  author={Zhang, Yumeng and Gong, Shi and Xiong, Kaixin and Ye, Xiaoqing and Li, Xiaofan and Tan, Xiao and Wang, Fan and Huang, Jizhou and Wu, Hua and Wang, Haifeng},
  journal={arXiv preprint arXiv:2407.05679},
  year={2024}
}

@article{wu2024holodrive,
  title={Holodrive: Holistic 2d-3d multi-modal street scene generation for autonomous driving},
  author={Wu, Zehuan and Ni, Jingcheng and Wang, Xiaodong and Guo, Yuxin and Chen, Rui and Lu, Lewei and Dai, Jifeng and Xiong, Yuwen},
  journal={arXiv preprint arXiv:2412.01407},
  year={2024}
}

@inproceedings{zyrianov2025lidardm,
  title={Lidardm: Generative lidar simulation in a generated world},
  author={Zyrianov, Vlas and Che, Henry and Liu, Zhijian and Wang, Shenlong},
  booktitle={2025 IEEE International Conference on Robotics and Automation (ICRA)},
  pages={6055--6062},
  year={2025},
  organization={IEEE}
}

@article{hu2023gaia,
  title={Gaia-1: A generative world model for autonomous driving},
  author={Hu, Anthony and Russell, Lloyd and Yeo, Hudson and Murez, Zak and Fedoseev, George and Kendall, Alex and Shotton, Jamie and Corrado, Gianluca},
  journal={arXiv preprint arXiv:2309.17080},
  year={2023}
}

@article{liao2025work,
  title={Work Zones challenge VLM Trajectory Planning: Toward Mitigation and Robust Autonomous Driving},
  author={Liao, Yifan and Sun, Zhen and Qiu, Xiaoyun and Zhao, Zixiao and Tang, Wenbing and He, Xinlei and Zheng, Xinhu and Zhang, Tianwei and Huang, Xinyi and Han, Xingshuo},
  journal={arXiv preprint arXiv:2510.02803},
  year={2025}
}

@inproceedings{radford2021learning,
  title={Learning transferable visual models from natural language supervision},
  author={Radford, Alec and Kim, Jong Wook and Hallacy, Chris and Ramesh, Aditya and Goh, Gabriel and Agarwal, Sandhini and Sastry, Girish and Askell, Amanda and Mishkin, Pamela and Clark, Jack and others},
  booktitle={International conference on machine learning},
  pages={8748--8763},
  year={2021},
  organization={PmLR}
}

@inproceedings{zhai2023sigmoid,
  title={Sigmoid loss for language image pre-training},
  author={Zhai, Xiaohua and Mustafa, Basil and Kolesnikov, Alexander and Beyer, Lucas},
  booktitle={Proceedings of the IEEE/CVF international conference on computer vision},
  pages={11975--11986},
  year={2023}
}

@article{cen2025worldvla,
  title={WorldVLA: Towards Autoregressive Action World Model},
  author={Cen, Jun and Yu, Chaohui and Yuan, Hangjie and Jiang, Yuming and Huang, Siteng and Guo, Jiayan and Li, Xin and Song, Yibing and Luo, Hao and Wang, Fan and others},
  journal={arXiv preprint arXiv:2506.21539},
  year={2025}
}

@inproceedings{zhao2025cot,
  title={Cot-vla: Visual chain-of-thought reasoning for vision-language-action models},
  author={Zhao, Qingqing and Lu, Yao and Kim, Moo Jin and Fu, Zipeng and Zhang, Zhuoyang and Wu, Yecheng and Li, Zhaoshuo and Ma, Qianli and Han, Song and Finn, Chelsea and others},
  booktitle={Proceedings of the Computer Vision and Pattern Recognition Conference},
  pages={1702--1713},
  year={2025}
}

@article{kim2024openvla,
  title={Openvla: An open-source vision-language-action model},
  author={Kim, Moo Jin and Pertsch, Karl and Karamcheti, Siddharth and Xiao, Ted and Balakrishna, Ashwin and Nair, Suraj and Rafailov, Rafael and Foster, Ethan and Lam, Grace and Sanketi, Pannag and others},
  journal={arXiv preprint arXiv:2406.09246},
  year={2024}
}

@article{li2025unified,
  title={Unified video action model},
  author={Li, Shuang and Gao, Yihuai and Sadigh, Dorsa and Song, Shuran},
  journal={arXiv preprint arXiv:2503.00200},
  year={2025}
}

@article{he2025pre,
  title={Pre-trained video generative models as world simulators},
  author={He, Haoran and Zhang, Yang and Lin, Liang and Xu, Zhongwen and Pan, Ling},
  journal={arXiv preprint arXiv:2502.07825},
  year={2025}
}

@article{jiang2023vad,
  title={VAD: Vectorized Scene Representation for Efficient Autonomous Driving},
  author={Jiang, Bo and Chen, Shaoyu and Xu, Qing and Liao, Bencheng and Chen, Jiajie and Zhou, Helong and Zhang, Qian and Liu, Wenyu and Huang, Chang and Wang, Xinggang},
  journal={ICCV},
  year={2023}
}

@inproceedings{hu2023_uniad,
 title={Planning-oriented Autonomous Driving}, 
 author={Yihan Hu and Jiazhi Yang and Li Chen and Keyu Li and Chonghao Sima and Xizhou Zhu and Siqi Chai and Senyao Du and Tianwei Lin and Wenhai Wang and Lewei Lu and Xiaosong Jia and Qiang Liu and Jifeng Dai and Yu Qiao and Hongyang Li},
 booktitle={Proceedings of the IEEE/CVF Conference on Computer Vision and Pattern Recognition},
 year={2023},
}

@inproceedings{dosovitskiy2017carla,
  title={CARLA: An open urban driving simulator},
  author={Dosovitskiy, Alexey and Ros, German and Codevilla, Felipe and Lopez, Antonio and Koltun, Vladlen},
  booktitle={Conference on robot learning},
  pages={1--16},
  year={2017},
  organization={PMLR}
}

@article{zeng2025futuresightdrive,
  title={FutureSightDrive: Thinking Visually with Spatio-Temporal CoT for Autonomous Driving},
  author={Zeng, Shuang and Chang, Xinyuan and Xie, Mengwei and Liu, Xinran and Bai, Yifan and Pan, Zheng and Xu, Mu and Wei, Xing},
  journal={arXiv preprint arXiv:2505.17685},
  year={2025}
}

@article{liu2025towards,
  title={Towards foundational LiDAR world models with efficient latent flow matching},
  author={Liu, Tianran and Zhao, Shengwen and Rhinehart, Nicholas},
  journal={arXiv preprint arXiv:2506.23434},
  year={2025}
}

@inproceedings{zheng2025world4drive,
  title={World4Drive: End-to-end autonomous driving via intention-aware physical latent world model},
  author={Zheng, Yupeng and Yang, Pengxuan and Xing, Zebin and Zhang, Qichao and Zheng, Yuhang and Gao, Yinfeng and Li, Pengfei and Zhang, Teng and Xia, Zhongpu and Jia, Peng and others},
  booktitle={Proceedings of the IEEE/CVF International Conference on Computer Vision},
  pages={28632--28642},
  year={2025}
}

@inproceedings{wang2025omnidrive,
  title={Omnidrive: A holistic vision-language dataset for autonomous driving with counterfactual reasoning},
  author={Wang, Shihao and Yu, Zhiding and Jiang, Xiaohui and Lan, Shiyi and Shi, Min and Chang, Nadine and Kautz, Jan and Li, Ying and Alvarez, Jose M},
  booktitle={Proceedings of the Computer Vision and Pattern Recognition Conference},
  pages={22442--22452},
  year={2025}
}

@article{fang2024eva,
  title={Eva-02: A visual representation for neon genesis},
  author={Fang, Yuxin and Sun, Quan and Wang, Xinggang and Huang, Tiejun and Wang, Xinlong and Cao, Yue},
  journal={Image and Vision Computing},
  volume={149},
  pages={105171},
  year={2024},
  publisher={Elsevier}
}

@article{wu2022trajectory,
  title={Trajectory-guided control prediction for end-to-end autonomous driving: A simple yet strong baseline},
  author={Wu, Penghao and Jia, Xiaosong and Chen, Li and Yan, Junchi and Li, Hongyang and Qiao, Yu},
  journal={Advances in Neural Information Processing Systems},
  volume={35},
  pages={6119--6132},
  year={2022}
}

@InProceedings{Jia_2023_CVPR,
    author    = {Jia, Xiaosong and Wu, Penghao and Chen, Li and Xie, Jiangwei and He, Conghui and Yan, Junchi and Li, Hongyang},
    title     = {Think Twice Before Driving: Towards Scalable Decoders for End-to-End Autonomous Driving},
    booktitle = {Proceedings of the IEEE/CVF Conference on Computer Vision and Pattern Recognition (CVPR)},
    month     = {June},
    year      = {2023},
    pages     = {21983-21994}
}

@inproceedings{jia2023driveadapter,
  title={DriveAdapter: Breaking the Coupling Barrier of Perception and Planning in End-to-End Autonomous Driving},
  author={Jia, Xiaosong and Gao, Yulu and Chen, Li and Yan, Junchi and Liu, Patrick Langechuan and Li, Hongyang},
  booktitle={ICCV},
  year={2023}
}

@article{zhai2023rethinking,
  title={Rethinking the open-loop evaluation of end-to-end autonomous driving in nuscenes},
  author={Zhai, Jiang-Tian and Feng, Ze and Du, Jinhao and Mao, Yongqiang and Liu, Jiang-Jiang and Tan, Zichang and Zhang, Yifu and Ye, Xiaoqing and Wang, Jingdong},
  journal={arXiv preprint arXiv:2305.10430},
  year={2023}
}

@inproceedings{zheng2024genad,
  title={Genad: Generative end-to-end autonomous driving},
  author={Zheng, Wenzhao and Song, Ruiqi and Guo, Xianda and Zhang, Chenming and Chen, Long},
  booktitle={European Conference on Computer Vision},
  pages={87--104},
  year={2024},
  organization={Springer}
}

@inproceedings{song2025don,
  title={Don't Shake the Wheel: Momentum-Aware Planning in End-to-End Autonomous Driving},
  author={Song, Ziying and Jia, Caiyan and Liu, Lin and Pan, Hongyu and Zhang, Yongchang and Wang, Junming and Zhang, Xingyu and Xu, Shaoqing and Yang, Lei and Luo, Yadan},
  booktitle={Proceedings of the Computer Vision and Pattern Recognition Conference},
  pages={22432--22441},
  year={2025}
}

@article{jia2025drivetransformer,
  title={Drivetransformer: Unified transformer for scalable end-to-end autonomous driving},
  author={Jia, Xiaosong and You, Junqi and Zhang, Zhiyuan and Yan, Junchi},
  journal={arXiv preprint arXiv:2503.07656},
  year={2025}
}

@inproceedings{kim2021drivegan,
  title={Drivegan: Towards a controllable high-quality neural simulation},
  author={Kim, Seung Wook and Philion, Jonah and Torralba, Antonio and Fidler, Sanja},
  booktitle={Proceedings of the IEEE/CVF Conference on Computer Vision and Pattern Recognition},
  pages={5820--5829},
  year={2021}
}

@article{wang2023drivedreamer,
  title={Drivedreamer: Towards real-world-driven world models for autonomous driving},
  author={Wang, Xiaofeng and Zhu, Zheng and Huang, Guan and Chen, Xinze and Zhu, Jiagang and Lu, Jiwen},
  journal={arXiv preprint arXiv:2309.09777},
  year={2023}
}

@article{wang2023driving,
  title={Driving into the Future: Multiview Visual Forecasting and Planning with World Model for Autonomous Driving},
  author={Wang, Yuqi and He, Jiawei and Fan, Lue and Li, Hongxin and Chen, Yuntao and Zhang, Zhaoxiang},
  journal={arXiv preprint arXiv:2311.17918},
  year={2023}
}

@article{doe,
    title={Doe-1: Closed-Loop Autonomous Driving with Large World Model},
    author={Zheng, Wenzhao and Xia, Zetian and Huang, Yuanhui and Zuo, Sicheng and Zhou, Jie and Lu, Jiwen},
    journal={arXiv preprint arXiv: 2412.09627},
    year={2024}
}

@article{li2022bevformer,
  title={BEVFormer: Learning Bird’s-Eye-View Representation from Multi-Camera Images via Spatiotemporal Transformers},
  author={Li, Zhiqi and Wang, Wenhai and Li, Hongyang and Xie, Enze and Sima, Chonghao and Lu, Tong and Qiao, Yu and Dai, Jifeng},
  journal={arXiv preprint arXiv:2203.17270},
  year={2022}
}

@article{chern2025thinking,
  title={Thinking with Generated Images},
  author={Chern, Ethan and Hu, Zhulin and Chern, Steffi and Kou, Siqi and Su, Jiadi and Ma, Yan and Deng, Zhijie and Liu, Pengfei},
  journal={arXiv preprint arXiv:2505.22525},
  year={2025}
}

@article{li2025imagine,
  title={Imagine while reasoning in space: Multimodal visualization-of-thought},
  author={Li, Chengzu and Wu, Wenshan and Zhang, Huanyu and Xia, Yan and Mao, Shaoguang and Dong, Li and Vuli{\'c}, Ivan and Wei, Furu},
  journal={arXiv preprint arXiv:2501.07542},
  year={2025}
}

@inproceedings{caesar2020nuscenes,
  title={nuscenes: A multimodal dataset for autonomous driving},
  author={Caesar, Holger and Bankiti, Varun and Lang, Alex H and Vora, Sourabh and Liong, Venice Erin and Xu, Qiang and Krishnan, Anush and Pan, Yu and Baldan, Giancarlo and Beijbom, Oscar},
  booktitle={Proceedings of the IEEE/CVF conference on computer vision and pattern recognition},
  pages={11621--11631},
  year={2020}
}

@inproceedings{sima2024drivelm,
  title={Drivelm: Driving with graph visual question answering},
  author={Sima, Chonghao and Renz, Katrin and Chitta, Kashyap and Chen, Li and Zhang, Hanxue and Xie, Chengen and Bei{\ss}wenger, Jens and Luo, Ping and Geiger, Andreas and Li, Hongyang},
  booktitle={European conference on computer vision},
  pages={256--274},
  year={2024},
  organization={Springer}
}

@article{cho2024language,
  title={Language-image models with 3d understanding},
  author={Cho, Jang Hyun and Ivanovic, Boris and Cao, Yulong and Schmerling, Edward and Wang, Yue and Weng, Xinshuo and Li, Boyi and You, Yurong and Kr{\"a}henb{\"u}hl, Philipp and Wang, Yan and others},
  journal={arXiv preprint arXiv:2405.03685},
  year={2024}
}

@article{ishaq2025tracking,
  title={Tracking meets large multimodal models for driving scenario understanding},
  author={Ishaq, Ayesha and Lahoud, Jean and Khan, Fahad Shahbaz and Khan, Salman and Cholakkal, Hisham and Anwer, Rao Muhammad},
  journal={arXiv preprint arXiv:2503.14498},
  year={2025}
}

@article{zheng2024simplellm4ad,
  title={Simplellm4ad: An end-to-end vision-language model with graph visual question answering for autonomous driving},
  author={Zheng, Peiru and Zhao, Yun and Gong, Zhan and Zhu, Hong and Wu, Shaohua},
  journal={arXiv preprint arXiv:2407.21293},
  year={2024}
}

@article{tian2024drivevlm,
  title={Drivevlm: The convergence of autonomous driving and large vision-language models},
  author={Tian, Xiaoyu and Gu, Junru and Li, Bailin and Liu, Yicheng and Wang, Yang and Zhao, Zhiyong and Zhan, Kun and Jia, Peng and Lang, Xianpeng and Zhao, Hang},
  journal={arXiv preprint arXiv:2402.12289},
  year={2024}
}

@article{li2025drivevla,
  title={DriveVLA-W0: World models amplify data scaling law in autonomous driving},
  author={Li, Yingyan and Shang, Shuyao and Liu, Weisong and Zhan, Bing and Wang, Haochen and Wang, Yuqi and Chen, Yuntao and Wang, Xiaoman and An, Yasong and Tang, Chufeng and others},
  journal={arXiv preprint arXiv:2510.12796},
  year={2025}
}

@article{dong2025unified,
  title={Unified world models: Memory-augmented planning and foresight for visual navigation},
  author={Dong, Yifei and Wu, Fengyi and Chen, Guangyu and Cheng, Zhi-Qi and Hu, Qiyu and Zhou, Yuxuan and Sun, Jingdong and He, Jun-Yan and Dai, Qi and Hauptmann, Alexander G},
  journal={arXiv preprint arXiv:2510.08713},
  year={2025}
}

@article{dong2026language,
  title={Language-Conditioned World Modeling for Visual Navigation},
  author={Dong, Yifei and Wu, Fengyi and Dai, Yilong and Kong, Lingdong and Chen, Guangyu and Zhu, Xu and Hu, Qiyu and Wang, Tianyu and Garnica, Johnalbert and Liu, Feng and others},
  journal={arXiv preprint arXiv:2603.26741},
  year={2026}
}

@article{chi2026driver,
  title={Driver-WM: A Driver-Centric Traffic-Conditioned Latent World Model for In-Cabin Dynamics Rollout},
  author={Chi, Haozhuang and Qiu, Daosheng and Su, Hao and Liu, Haochen and Li, Zirui and Zhang, Haoruo and Lv, Chen},
  journal={arXiv preprint arXiv:2605.05092},
  year={2026}
}

@article{yu2026g2dp,
  title={G2DP: Diffusion Planning with Spatio-Temporal Grid Guidance},
  author={Yu, Hang and Jin, Ye and Canevaro, Alessandro and Schmidt, Julian and Jordan, Julian and Li, Peizheng and Kaufeld, Marc and Lindner, Silvan and Betz, Johannes and Stork, Wilhelm},
  journal={arXiv preprint arXiv:2606.26017},
  year={2026}
}

@article{wei2025lidardraft,
  title={LiDARDraft: Generating LiDAR Point Cloud from Versatile Inputs},
  author={Wei, Haiyun and Lu, Fan and Zhu, Yunwei and Zheng, Zehan and Xue, Weiyi and Shao, Lin and Zhang, Xudong and Wu, Ya and Fu, Rong and Chen, Guang},
  journal={arXiv preprint arXiv:2512.20105},
  year={2025}
}

@article{liu2026towards,
  title={Towards foundational LiDAR world models with efficient latent flow matching},
  author={Liu, Tianran and Zhao, Shengwen and Rhinehart, Nicholas},
  journal={Advances in Neural Information Processing Systems},
  volume={38},
  pages={155959--155994},
  year={2026}
}

@article{liu2026occsim,
  title={OccSim: Multi-kilometer Simulation with Long-horizon Occupancy World Models},
  author={Liu, Tianran and Zhao, Shengwen and Pourkeshavarz, Mozhgan and Li, Weican and Rhinehart, Nicholas},
  journal={arXiv preprint arXiv:2603.28887},
  year={2026}
}

@inproceedings{han2025extrapolated,
  title={Extrapolated urban view synthesis benchmark},
  author={Han, Xiangyu and Jia, Zhen and Li, Boyi and Wang, Yan and Ivanovic, Boris and You, Yurong and Liu, Lingjie and Wang, Yue and Pavone, Marco and Feng, Chen and others},
  booktitle={Proceedings of the IEEE/CVF International Conference on Computer Vision},
  pages={28718--28728},
  year={2025}
}

@article{liu2026driveva,
  title={Driveva: Video action models are zero-shot drivers},
  author={Liu, Mengmeng and Zhang, Diankun and Liu, Jiuming and Cui, Jianfeng and Xie, Hongwei and Chen, Guang and Ye, Hangjun and Yang, Michael Ying and Nex, Francesco and Cheng, Hao},
  journal={arXiv preprint arXiv:2604.04198},
  year={2026}
}

@article{yao2026vlga,
  title={VLGA: Vision-Language-Geometry-Action Models for Autonomous Driving},
  author={Yao, Jin and Kurra, Dhruva Dixith and Lampo, Tom and Cheng, Zezhou and Guo, Danhua and Yaman, Burhan},
  journal={arXiv preprint arXiv:2606.12396},
  year={2026}
}

% \maketitlesupplementary

\clearpage
\appendix
\section*{\huge Appendix}
% \addcontentsline{toc}{section}{Appendix} % 可选，arxiv一般无所谓

\begin{table*}[ht!]
\centering
\caption{Multi-ability results on the Bench2Drive base set. An asterisk denotes expert feature distillation. C/L refers to camera/LiDAR. NC: navigation command; TP: target point.}
\small
\label{tab:multi-ability}
\resizebox{\textwidth}{!}{%
\begin{tabular}{lcccccccc}
\specialrule{1.2pt}{0pt}{0pt}
\multirow{2}{*}{Method} &
\multirow{2}{*}{Condition} &
\multirow{2}{*}{Modality} &
\multicolumn{6}{c}{Ability (\%) $\uparrow$} \\

\cmidrule(lr){4-9}
 &  &  & Merging & Overtaking & Emergency Brake & Give Way & Traffic Sign & Mean \\
\midrule
TCP*~\citep{wu2022trajectory} & TP & C & 16.18 & 20.00 & 20.00 & 10.00 & 6.99 & 14.63 \\
TCP-ctrl* & TP & C & 10.29 & 4.44 & 10.00 & 10.00 & 6.45 & 8.23 \\
TCP-traj* & TP & C & 8.89 & 24.29 & 51.67 & 40.00 & 46.28 & 34.22 \\
TCP-traj w/o distillation & TP & C & 17.14 & 6.67 & 40.00 & 50.00 & 28.72 & 28.51 \\
ThinkTwice*~\citep{Jia_2023_CVPR} & TP & C & 27.38 & 18.42 & 35.82 & 50.00 & 54.23 & 37.17 \\
DriveAdapter*~\citep{jia2023driveadapter} & TP & C\&L & 28.82 & 26.38 & 48.76 & 50.00 & 56.43 & 42.08 \\
\midrule
AD-MLP~\citep{zhai2023rethinking} & NC & C & 0.00 & 0.00 & 0.00 & 0.00 & 4.35 & 0.87 \\
UniAD-Tiny~\citep{hu2023_uniad} & NC & C & 8.89 & 9.33 & 20.00 & 20.00 & 15.43 & 14.73 \\
UniAD-Base~\citep{hu2023_uniad} & NC & C & 14.10 & 17.78 & 21.67 & 10.00 & 14.21 & 15.55 \\
VAD~\citep{jiang2023vad} & NC & C & 8.11 & 24.44 & 18.64 & 20.00 & 19.15 & 18.07 \\
DriveTransformer-Large~\citep{jia2025drivetransformer} & NC & C & 17.57 & 35.00 & 48.36 & 40.00 & 52.10 & 38.60 \\
\textbf{ORION} & NC & C & 25.00 & 71.11 & 78.33 & 30.00 & 69.15 & 54.72 \\
\midrule
\textbf{Ours(AR)} & NC & C & \textbf{29.81} & \textbf{74.04} & \textbf{79.84} & 40.00 & \textbf{71.30} & \textbf{59.00} \\
\textbf{Ours(AR+Diffusion)} & NC & C & \textbf{29.97} & \textbf{74.64} & \textbf{79.98} & 40.00 & \textbf{71.54} & \textbf{59.23} \\
\specialrule{1.2pt}{0pt}{0pt}
\end{tabular}
}
\end{table*}

\section{Evaluation Metrics}

For closed-loop evaluation, we follow the evaluation pipeline of Bench2Drive~\citep{jia2024bench2drive}, which adopts four key metrics: Driving Score (DS), Success Rate (SR), Efficiency, and Comfortness. Success Rate measures the percentage of routes that are successfully completed within the designated time limit. Driving Score follows the CARLA~\cite{dosovitskiy2017carla} protocol and combines route completion with violation penalties, where infractions proportionally reduce the score through discount factors. Efficiency and Comfortness assess the speed performance and ride smoothness, respectively, of the autonomous driving policy throughout the episode.
% Multi-Ability evaluates five advanced driving skills independently, providing a comprehensive assessment of urban-driving competence.

For open-loop evaluation, we report L2 trajectory error and collision rate. We also evaluate 3D detection performance using standard nuScenes metrics, including mAP, mATE, mASE, mAOE, mAVE, and NDS, which measure detection accuracy, translation error, scale error, orientation error, velocity error, and overall detection quality, respectively. For future frame prediction, we use the Fréchet Inception Distance (FID) to evaluate how closely the distribution of generated frames matches that of the ground-truth future frames.

For VQA evaluation, we adopt the DriveLM GVQA~\citep{sima2024drivelm} metrics, including BLEU, ROUGE-L, and CIDEr for language generation, ChatGPT score for open-ended question answering, and accuracy for multiple-choice questions.

\section{More Results}

\subsection{Additional Closed-Loop Evaluation Results}

We provide additional closed-loop multi-ability evaluation results on Bench2Drive in Tab.~\ref{tab:multi-ability}.
Compared with baseline methods, our approach achieves stronger performance in Merging, Overtaking, Emergency Brake, and Traffic Sign tasks. These gains suggest that leveraging the VLM's reasoning capability together with future-image prediction improves overall multi-ability performance, as reflected in the higher mean score.

\subsection{Further Visualization}

\begin{figure*}[ht!]
    \centering
    \includegraphics[width=\linewidth]{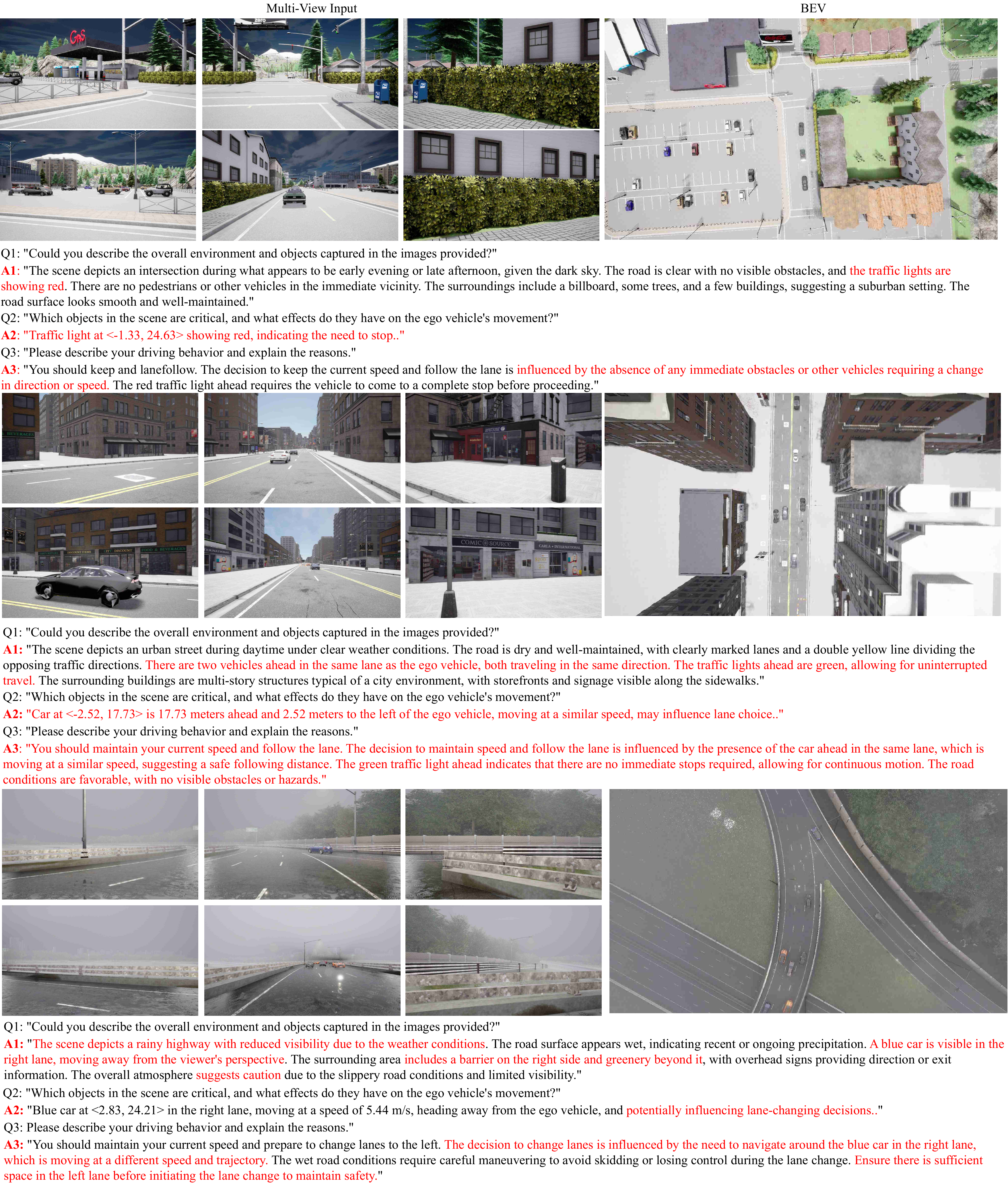}
    \caption{Visualization of our VQA results on Bench2Drive open-loop evaluation. For each scene, we show the multi-view input images and the bird's-eye view (BEV). The question-answer pair includes the model's understanding of the current state and reasoning about the ego vehicle's movement and driving behavior. Red text highlights our method's scene comprehension and reasoning.}
    
    \label{fig:supp_vqa}
\end{figure*}

We present visualizations of the VQA results from open-loop evaluation in Fig.~\ref{fig:supp_vqa}. These examples show that our model not only understands the current scene but also reasons effectively about future states, enabling reliable, decision-oriented responses even in complex driving scenarios and challenging weather conditions.

\begin{figure*}[ht!]
    \centering
    \includegraphics[width=0.66\linewidth]{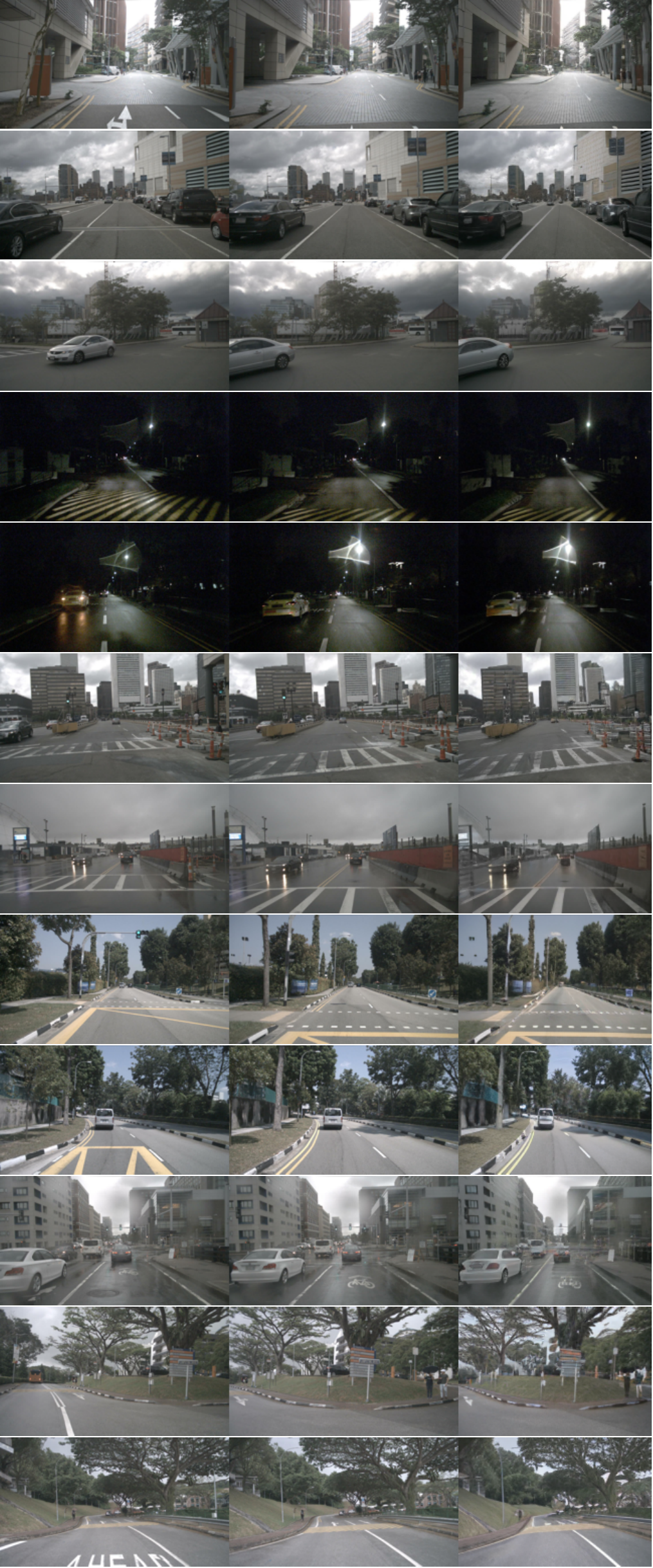}
    \caption{Qualitative results on nuScenes using the AR architecture.}
    
    \label{fig:supp_nuscenes_AR}
\end{figure*}

\begin{figure*}[ht!]
    \centering
    \includegraphics[width=0.7\linewidth]{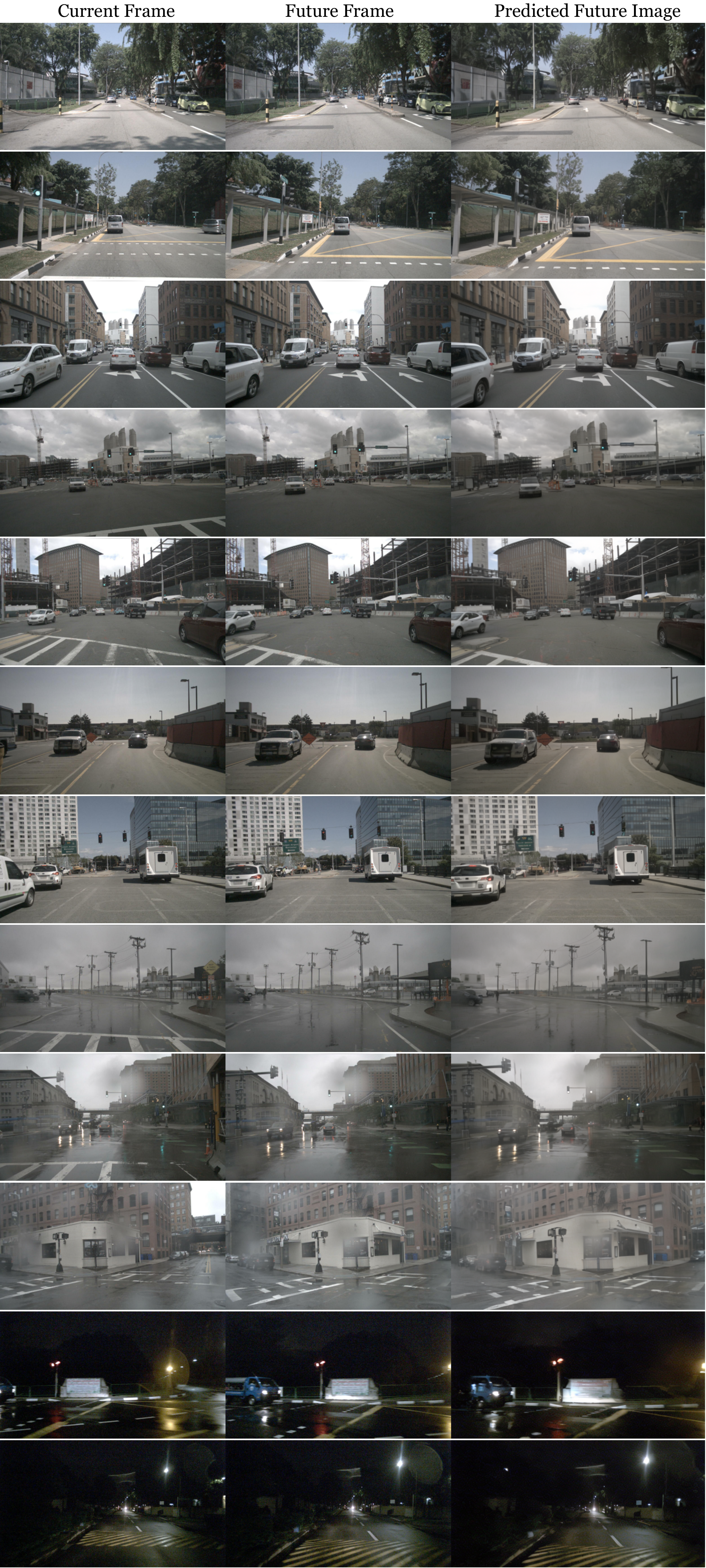}
    \caption{Qualitative results on nuScenes using the AR+Diffusion architecture.}
    
    \label{fig:supp_nuscenes_AR+Diff}
\end{figure*}

We further provide qualitative results for our AR and AR+Diffusion architectures in Fig.~\ref{fig:supp_nuscenes_AR} and Fig.~\ref{fig:supp_nuscenes_AR+Diff}, respectively. These visualizations indicate that both architectures can generate plausible future frames across diverse scenes and weather conditions.

\section{Additional Details on Perception Module}

\subsection{QT-Former Backbone}

The perception module in UniDrive-WM builds on the QT-Former perception backbone~\citep{fu2025orion}. 
As described in the main paper, QT-Former encodes multi-view observations and temporal history into structured visual representations. 
Specifically, we initialize learnable scene queries $Q_s \in \mathbb{R}^{N_s \times C_q}$ and perception queries $Q_p \in \mathbb{R}^{N_p \times C_q}$, where $N_s$ and $N_p$ denote the numbers of scene and perception queries, and $C_q$ is the query dimension. 
These queries interact with the current multi-view image features $F_m$ and their associated 3D positional encodings through cross-attention to compress the visual observations into compact scene-aware representations. 
In addition, QT-Former maintains a set of history queries $Q_h \in \mathbb{R}^{N_h \times C_q}$ together with a long-term memory bank $M \in \mathbb{R}^{(N_h \times n)\times C_q}$ to retrieve temporally relevant information from previous frames. 
The updated history queries and current scene features are further projected into the LLM reasoning space, while the perception queries $Q_p$ are connected to task-specific perception heads.

\subsection{Detection Heads and Supervision}

Concretely, the perception branch contains multiple auxiliary multilayer perceptron (MLP) heads for 3D object detection (including critical objects and lanes), as well as traffic-state estimation and motion prediction for dynamic agents.
Among these heads, only the detection head is used to report the 3D detection metrics (e.g., mAP and NDS) in the main paper.
The remaining heads are retained as auxiliary supervision to improve the quality of the shared perception features and preserve traffic-aware scene structure in the learned QT-Former representations.

Following ORION, the perception branch is pretrained as a set prediction problem with explicit supervision on these tasks. 
In particular, optimal query-to-object assignments are established via bipartite matching using the Hungarian algorithm. 
The 3D detection objective then consists of a focal loss for classification and an $\ell_1$ loss for continuous bounding box regression (center coordinates, dimensions, yaw, and velocity). 
The traffic-state branch is supervised by a focal loss, and the motion-prediction branch is trained with a focal loss together with an $\ell_1$ regression loss. 
These objectives provide explicit object-aware, traffic-aware, and motion-aware supervision for the QT-Former perception backbone.

From the perspective of UniDrive-WM, the role of the perception module is not limited to box prediction. 
More importantly, it acts as a structured perception encoder that provides reliable visual grounding for downstream reasoning, trajectory planning, and future image generation. 
By maintaining explicit supervision on critical traffic elements and dynamic cues, the perception backbone supplies the VLM with more informative scene representations for unified world modeling.

During joint planning and image generation training in UniDrive-WM, we initialize the model with the pretrained QT-Former and freeze the detection head.
Therefore, the joint optimization involves only the downstream planning and image-generation objectives, while the perception module mainly serves as a stable source of structured perception features rather than being jointly optimized as an end task.
This design is also consistent with our ablation results, where disabling detection supervision degrades planning performance, indicating that explicit perception learning improves the quality of the shared representations and remains important for reliable driving behavior.

\section{Additional Details on Trajectory Planner}

The trajectory planner in UniDrive-WM serves as a differentiable bridge between the semantic reasoning space of the VLM and the continuous action space of future trajectories. Given the current multimodal state representation $\mathbf{s}_t$ and the high-level reasoning embedding $\mathbf{h}_{\text{VLM}}$ produced by the VLM, the planner models a conditional distribution over future waypoints:
\begin{equation}
p_\theta(\mathbf{a}_{t:t+m} \mid \mathbf{s}_t, \mathbf{h}_{\text{VLM}}),
\end{equation}
where $\mathbf{a}_{t:t+m}$ denotes the predicted future trajectory over the planning horizon.

Following the generative-planning formulation in ORION, we introduce a latent variable to bridge the gap between the reasoning space and the action space. Specifically, the planner first projects the VLM reasoning embedding into a Gaussian latent space using lightweight MLP layers:
\begin{equation}
q_\phi(\mathbf{z}_a \mid \mathbf{h}_{\text{VLM}}) = \mathcal{N}(\boldsymbol{\mu}_h, \boldsymbol{\sigma}_h^2),
\end{equation}
where $\mathbf{z}_a$ denotes the latent action variable, and the mean $\boldsymbol{\mu}_h$ and variance $\boldsymbol{\sigma}_h^2$ are predicted from the planning-related hidden representation. To allow backpropagation through the stochastic sampling process, we employ the reparameterization trick: $\mathbf{z}_a = \boldsymbol{\mu}_h + \boldsymbol{\sigma}_h \odot \boldsymbol{\epsilon}$, where $\boldsymbol{\epsilon} \sim \mathcal{N}(0, \mathbf{I})$. 

The latent code is then used to decode the future ego trajectory. In practice, following ORION, we use a lightweight recurrent waypoint decoder. This decoder directly regresses the future trajectory represented as a sequence of $m$ continuous 2D waypoints in the ego vehicle's bird's-eye-view (BEV) coordinate system, i.e., $\hat{\mathbf{a}}_{t:t+m} = \{ (\hat{x}_i, \hat{y}_i) \}_{i=1}^m$. Compared with directly regressing waypoints from language features, this latent-variable design provides a smoother and more differentiable interface between reasoning and action. This allows the planner to preserve high-level semantic intent from the VLM while producing numerically precise trajectories in continuous space.

Unlike the original ORION formulation, which explicitly aligns the reasoning token and the ground-truth trajectory in a Gaussian latent space with a KL-divergence objective, we omit the explicit KL regularization term during our joint multimodal training. We empirically found that forcing the latent space to conform strictly to a standard normal distribution can destabilize the multimodal training process in large-scale, VLM-centric setups. Instead, we optimize the planning head using the composite objective defined in the main paper:
\begin{equation}
\mathcal{L}_{\text{plan}} = \mathcal{L}_{\text{col}} + \mathcal{L}_{\text{bd}} + \mathcal{L}_{\text{mse}},
\end{equation}
where the losses provide explicit, spatially aware supervision to ensure physical safety and perception-action coupling:

\begin{itemize}
    
    \item \textbf{Collision Loss ($\mathcal{L}_{\text{col}}$):} To explicitly couple the action space with the perception outputs, $\mathcal{L}_{\text{col}}$ penalizes trajectories that intersect with the predicted 3D bounding boxes of dynamic agents (supplied by the frozen detection head). This ensures that the planner respects the occupancy of surrounding traffic.
    \item \textbf{Boundary Loss ($\mathcal{L}_{\text{bd}}$):} Similarly, $\mathcal{L}_{\text{bd}}$ penalizes trajectory points that fall outside the drivable area or cross the predicted solid lane boundaries, enforcing adherence to road topology.
    \item \textbf{MSE Loss ($\mathcal{L}_{\text{mse}}$):} This term is the standard $\ell_2$ distance between the predicted waypoints $\{ (\hat{x}_i, \hat{y}_i) \}_{i=1}^m$ and the ground-truth expert trajectory. It provides the primary driving demonstration signal.
\end{itemize}

From the perspective of UniDrive-WM, the trajectory planner is not merely an output head for motion prediction. Rather, it is the key module that connects reasoning and visual imagination: the planned trajectory conditions future image generation, and the generated future frames in turn provide additional supervisory signals that improve planning quality. This is supported by our ablation results, where conditioning image generation on the planning trajectory yields better planning accuracy and visual quality, aligning with the causal direction from action to future observation.

\end{document}